\documentclass{article} 
\usepackage{iclr2021_conference,times}

\usepackage{amsmath,amsfonts,bm}

\def\eqref#1{equation~\ref{#1}}

\def\1{\bm{1}}

\DeclareMathAlphabet{\mathsfit}{\encodingdefault}{\sfdefault}{m}{sl}
\SetMathAlphabet{\mathsfit}{bold}{\encodingdefault}{\sfdefault}{bx}{n}

\usepackage{graphicx}
\usepackage{graphics}
\usepackage{hyperref}
\usepackage{url}
\usepackage{multirow}

\usepackage{colortbl}
\usepackage{makecell}

\usepackage{booktabs}
\usepackage{arydshln}

\def\best{\bf\cellcolor[gray]{0.85}}
\def\secbest{\cellcolor[gray]{0.92}}
\definecolor{TableRed}{HTML}{800000}
\definecolor{mediumelectricblue}{rgb}{0.01, 0.31, 0.59}

\newcommand{\electricblue}[1]{\textcolor{mediumelectricblue}{#1}}

\usepackage{algorithm}
\usepackage{algorithmicx}
\usepackage{algpseudocode}

\usepackage{color}
\definecolor{org}{rgb}{0.992, 0.753, 0.525}

\iclrfinalcopy

\newif\ifsubmit
\ifsubmit
\newcommand{\xiao}[1]{}
\newcommand{\dawn}[1]{}
\else
\newcommand{\xiao}[1]{\textcolor{green}{[Xiao: #1]}}
\newcommand{\dawn}[1]{\textcolor{red}{[Dawn: #1]}}
\fi
\submitfalse

\newcommand{\texttu}[1]{\texttt{\uppercase{#1}}}
\newcommand{\method}{\textsc{MaMa}}
\newcommand{\stepone}{Match}
\newcommand{\steptwo}{Map}
\newcommand{\hide}[1]{}

\usepackage{mdwlist}
\usepackage[list=true]{subcaption}
\makeatletter
\newcommand\footnoteref[1]{\protected@xdef\@thefnmark{\ref{#1}}\@footnotemark}
\makeatother

\title{Language Models are \\Open Knowledge Graphs}

\author{Chenguang Wang$^*$, Xiao Liu$^\P$, Dawn Song$^*$ \\
$^*$UC Berkeley\\
$^\P$Tsinghua University \\
\texttt{\{chenguangwang,dawnsong\}@berkeley.edu} \\
\texttt{liuxiao17@mails.tsinghua.edu.cn} \\
}

\begin{document}

\maketitle

\begin{abstract}
This paper shows how to construct knowledge graphs (KGs) from pre-trained language models (e.g., BERT, GPT-2/3), without human supervision.
Popular KGs (e.g, Wikidata, NELL) are built in either a supervised or semi-supervised manner, requiring humans to create knowledge. Recent deep language models automatically acquire knowledge from large-scale corpora via pre-training. The stored knowledge has enabled the language models to improve downstream NLP tasks, e.g., answering questions, and writing code and articles. In this paper, we propose an unsupervised method to cast the knowledge contained within language models into KGs. We show that KGs are constructed with a single forward pass of the pre-trained language models (without fine-tuning) over the corpora. We demonstrate the quality of the constructed KGs by comparing to two KGs (Wikidata, TAC KBP) created by humans. Our KGs also provide open factual knowledge that is new in the existing KGs. Our code and KGs will be made publicly available.
\end{abstract}
\section{Introduction}
Knowledge graphs (KGs) are an important resource for both humans and machines. Factual knowledge in KGs is injected into AI applications to imitate important skills possessed by humans, e.g., reasoning and understanding. KG construction is mainly supervised, requiring humans to handwrite every fact, such as Freebase~\citep{bollacker2008freebase} and Wikidata. KGs can also be constructed in a semi-supervised way, in which a semi-automatic extractor is used to obtain the facts from web corpora (e.g., NELL~\citep{carlson2010toward} and Knowledge Vault~\citep{dong2014knowledge}). Humans however still need to interact with the extractor to improve the quality of the discovered facts. Therefore, human supervision, which is often expensive, is required in constructing KGs.

Recent progress in language models (LMs), such as BERT~\citep{devlin2018bert} and GPT-2/3~\citep{radford2019language,brown2020language}, has led to superior results even outperforming humans in a wide range of tasks, e.g., sentence classification~\citep{wang2018glue}, question answering~\citep{brown2020language}. Pre-trained LMs are also capable to write poetry, music, and code, while such tasks often require we human to spend a significant amount of time in learning the relevant knowledge to work well. In fact, these pre-trained LMs automatically acquire factual knowledge from large-scale corpora (e.g., BookCorpus~\citep{zhu2015aligning}, Common Crawl~\citep{brown2020language}) via pre-training. The learned knowledge in pre-trained LMs is the key to the current success. We therefore consider the following question: instead of using the manually created knowledge, \emph{can we use the knowledge stored in pre-trained LMs to construct KGs?}

\begin{figure*}
    \centering
    \includegraphics[width=1.0\textwidth]{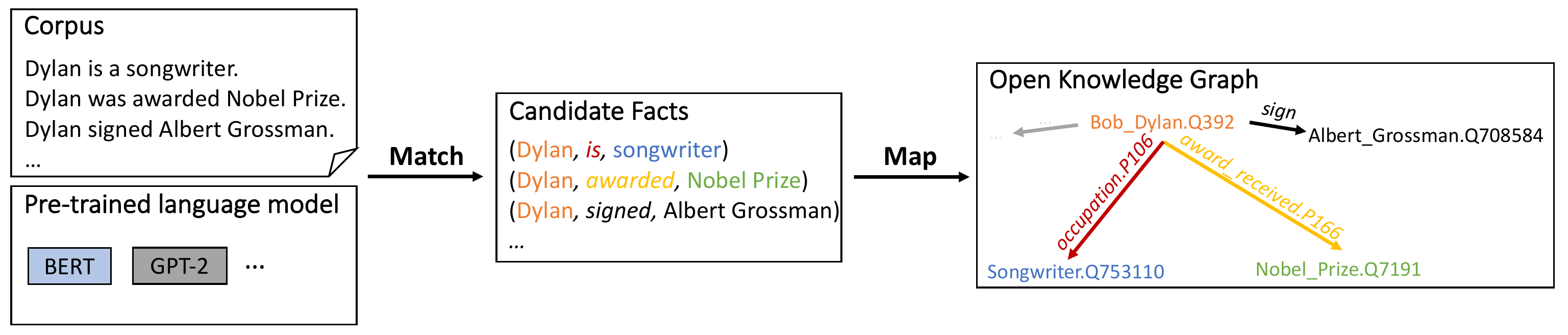}
    \vspace{-0.05in}
    \caption{{\small Overview of the proposed approach \method. \method\ constructs an open knowledge graph (KG) with a single forward pass of the pre-trained language model (LM) (without fine-tuning) over the corpus. Given the input: a textual corpus containing passages and sentences, e.g., English Wikipedia, and a pre-trained LM, e.g., BERT, GPT-2/3, \method\ (1) generates a set of candidate facts via {\it matching} the knowledge in the pre-trained LM with facts in the textual corpus, e.g., a candidate fact {\sl (Dylan, is, songwriter)} from the sentence {``Dylan is a songwriter.''}, and (2) produces an open KG by {\it mapping} the matched candidate facts to both an existing KG schema, e.g., {\sl (Bob\_Dylan.Q392, occupation.P106, Songwriter.Q753110)} in Wikidata schema, and an open schema, e.g., {\sl (Bob\_Dylan.Q392, sign, Albert\_Grossman.Q708584)}.}
      \label{fig:overview}}
\end{figure*}

In this paper, we design an unsupervised approach called \method\ that successfully recovers the factual knowledge stored in LMs to build KGs from scratch. \method\ constructs a KG with a single forward pass of a pre-trained LM (without fine-tuning) over a textual corpus. As illustrated in Figure~\ref{fig:overview}, \method\ has two stages: \stepone\ and \steptwo. 
{\bf \stepone}\ stage generates a set of candidate facts by matching the facts in the textual corpus with the knowledge in the pre-trained LM. General or world knowledge from large-scale corpora is embedded in the LM, thus candidate facts in the target corpus are often covered by the knowledge in the LM. The candidate facts are matched through an efficient beam search in the attention weight matrices of the pre-trained LM without fine-tuning. 
{\bf \steptwo}\ stage produces an open KG via mapping the matched candidate facts from \stepone\ stage to both fixed KG schema and open schema. If the schema of candidate facts exists in the KG schema, we map the candidate facts directly to the fixed KG schema. Otherwise, we reserve the unmapped candidate facts in the open schema. 
This results in a new type of KG, {\em open KG}, with a mixture of mapped facts in fixed KG schema and unmapped facts in the open schema.

Our contributions are as follows:
\begin{enumerate*}
\vspace{-0.12in}
    \item We show {\it how to construct KGs from pre-trained LMs}. The KGs are constructed with a single forward pass of the pre-trained LMs without fine-tuning over the textual corpora. This helps researchers explicitly understand what the language models learn, bridging the deep LM and KG communities through enhanced model transparency.
    \item We propose an unsupervised two-stage approach, \method, to first {\em match} the candidate facts in the corpora with the knowledge stored in LMs, then {\em map} the matched candidate facts to both fixed and open schema to produce a KG.
    \item We generate a new type of KG, namely open KG, consists of {\em mapped facts} in the fixed KG schema of existing KGs (Wikidata and TAC KBP) annotated by humans; and {\it unmapped facts} in the open schema that are new in the reference KG schema. The reach of this result is broad and has downstream utility for knowledge graph construction, deep neural network interpretation, and information extraction.
\end{enumerate*}

\vspace{-0.05in}
\section{\method}
\label{sec:app}
\vspace{-0.05in}

We introduce an unsupervised end-to-end approach \stepone\ and \steptwo\ (\method) as illustrated in Figure~\ref{fig:overview} to construct open knowledge graphs (KGs) from language models (LMs). \method\ constructs the KGs with a single forward pass of the pre-trained LMs (without fine-tuning) over the corpora. The two stages of \method\ are:

\vspace{-0.05in}
{\bfseries \stepone} generates a set of candidate facts from a textual corpus. LMs contain global or world knowledge learned from large-scale corpora, which often does not perfectly match the knowledge in the target corpus. The goal of this stage is to match the knowledge stored in pre-trained LMs with facts in the corpus. Each fact is represented as a triplet {\sl (head, relation, tail)}~\footnote{\tiny{We use the term ``head'' and ``tail'' to denote head and tail's ``entities'' or ``entity mentions'' for simplicity.}}, in short, $(h, r, t)$, and passed to \steptwo\ stage. \stepone\ procedure is detailed in Sec.~\ref{sec:search}.

\vspace{-0.05in}
{\bfseries \steptwo} produces an open KG using the matched candidate facts from \stepone\ stage. The constructed open KG has two portions: (a) mapped candidate facts that are in a fixed KG schema, e.g., {\sl (Dylan, is, songwriter)} is mapped to {\sl (Bob\_Dylan.Q392, occupation.P106, Songwriter.Q753110)} according to Wikidata schema; and (b) unmapped candidate facts that are in an open schema, e.g., a candidate fact {\sl (Dylan, signed, Albert Grossman)} is partially mapped to {\sl (Bob\_Dylan.Q392, sign, Albert\_Grossman.Q708584)} in the open schema. This stage is described in Sec.~\ref{sec:kd}.

\vspace{-0.05in}
\subsection{\stepone}
\label{sec:search}
\vspace{-0.05in}

\begin{figure*}
\centering
\subcaptionbox{{\small Matching example.}}{\includegraphics[width=0.45\textwidth]{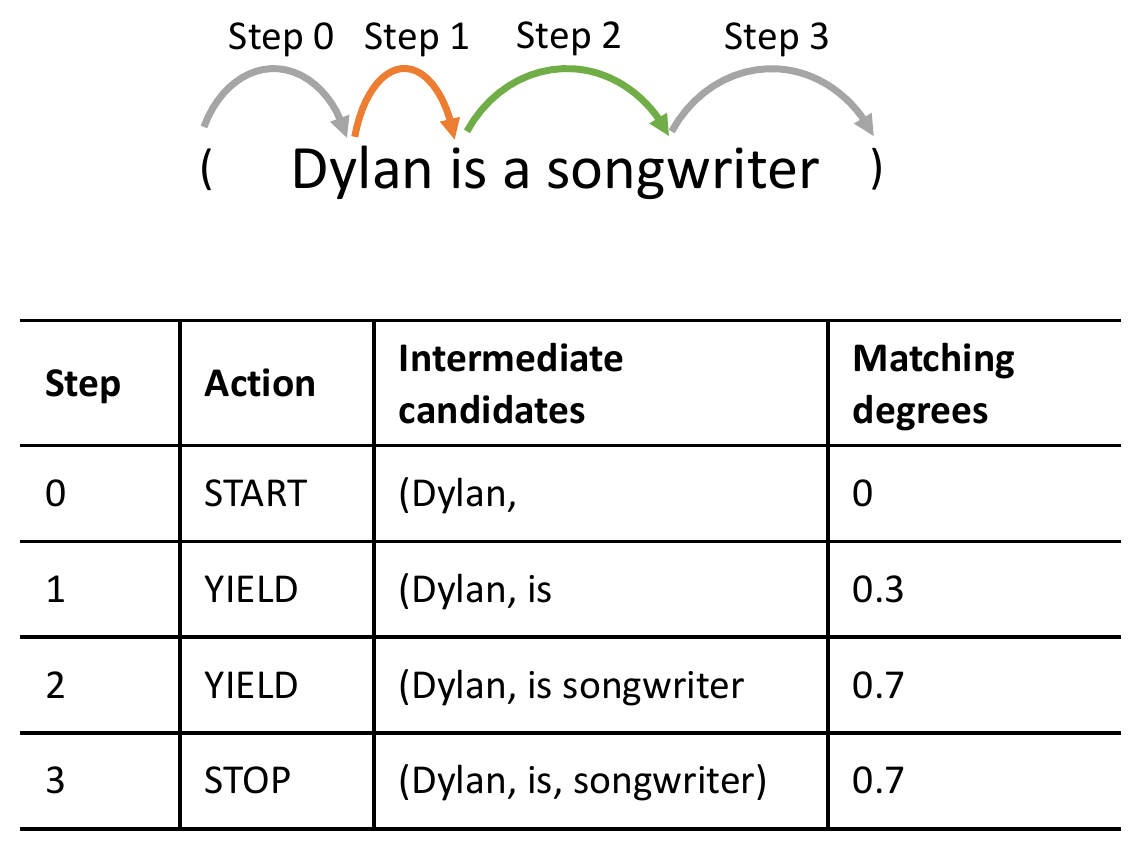}}%
\hspace{0.2in}
\subcaptionbox{{\small Attention matrix for matching degree calculation.}}{\includegraphics[width=0.33\textwidth]{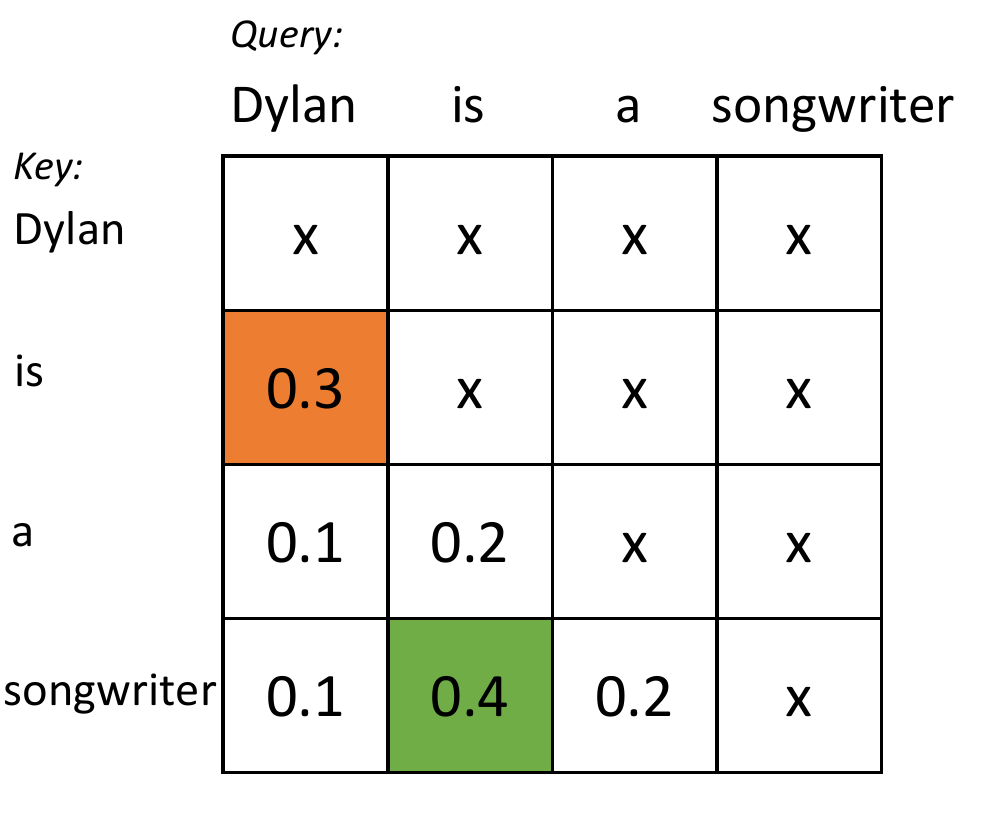}}%
\caption{{\small Illustration of \stepone\ stage. The upper part of (a) represents the general matching steps of generating the best matched candidate fact {\sl (Dylan, is, songwriter)} from the sentence ``Dylan is a songwriter.'' The lower portion shows the corresponding step-by-step process. Given a head-tail pair {\sl (Dylan, songwriter)}, at each step, the search chooses one of the actions, i.e., {\texttu{start}, \texttu{yield}, \texttu{stop}} to produce an intermediate candidate fact. The search {\em starts} by adding the head ``Dylan'' as an initial candidate (step 0). The matching degree of the candidate is initialized as 0. Next, a new candidate is {\em yielded} if the candidate has not reached the tail ``songwriter'', by appending the next largest attended token (with the largest score from the attention matrix (b) of the sentence) to the end of the current candidate, and the corresponding matching degrees are increased by the associated attention scores (step 1 and step 2). Otherwise, the search {\em stops}, and the candidate fact with the best matching degree is returned for the head-tail pair (step 3). The attention matrix (b) is from the forward pass of the LM without fine-tuning over the sentence. ``x'' marks the tokens to prevent searching backward.}}
\label{fig:search}
\end{figure*}


We frame the matching procedure as a search problem. To obtain the best matched candidate facts of an input sentence, the candidates with the top matching degrees are returned from a search process. The matching degree is derived from the search in the attention weight matrices of the pre-trained LM, since the attention weight matrices are the container of the knowledge in the pre-trained LM. The attention weight matrices are simply from the forward pass of the LM without fine-tuning over the sentence.


\vspace{-0.05in}
\subsubsection{Beam Search}
\vspace{-0.05in}


We design a simple-yet-effective beam search to find the best matched candidate facts. For every head-tail pair $(h, t)$ in a sentence, the search maintains the $k$-best matched candidate facts of the pair. Let's first consider the search from left to right with beam size equals to 1. An example search process is shown in Figure~\ref{fig:search}. Given a head-tail pair {\sl (Dylan, songwriter)}, at each step, the search performs one of the following actions:
\begin{description*}
\vspace{-0.1in}
\item [\texttu{start}] the search from the head. The head $h$ is added as an initial candidate into the beam. For simplicity, we use $\texttu{start}(h)$ to denote the action, which returns a candidate $(h,$. In Figure~\ref{fig:search}(a), at step 0, the head ``Dylan'' is added as {\sl (Dylan, } into the beam. The matching degree is initialized to 0.
\item [\texttu{yield}] a new intermediate candidate in the beam if the current candidate has not reached the tail. The next largest attended token (with the largest score from the attention matrix) is appended to the end of the current candidate to yield the new candidate. The corresponding matching degrees are increased by the associated attention scores. At step 1 (orange arrow in Figure~\ref{fig:search}(a)), ``is'' is appended to the current candidate to yield {\sl (Dylan, is, }, since ``is'' has the largest attention score with ``Dylan'' in the attention matrix. The attention score is 0.3 as highlighted in orange in Figure~\ref{fig:search}(b). The multi-head attention is reduced to a single head so that every two tokens of the sentence are associated with one attention weight. We experiment with different reduction setups in Sec.~\ref{sec:para}. ``x'' marks the tokens (prior to the current token) that are not considered in the search to prevent searching backward. Step 2 similarly takes $\texttu{yield}$ action to produce {\sl (Dylan, is songwriter, }. We use $\texttu{yield}(c,s,\mathbf{A}_s)$ to denote the action, where $c$ is a current candidate, $s$ represents the sentence, and $\mathbf{A}_s$ is the attention matrix from the forward pass of the pre-trained LM over $s$, which yields a new candidate.
\item [\texttu{stop}] the search step if the candidate has reached the tail, then add the candidate as a valid candidate fact into the beam. As beam size equals to 1, {\sl (Dylan, is, songwriter)} is the only returned candidate fact for the given pair. We denote this step using $\texttu{stop}(c,t)$, which returns a valid fact.
\end{description*}

\vspace{-0.05in}
The details of the proposed beam search are in Algorithm~\ref{alg:beamsearch}. The inputs of the search algorithm are a head-tail pair $(h,t)$, a sentence $s$, an attention matrix $\mathbf{A}_s$ of $s$. Both $h$ and $t$ are identified by the noun chunk in $s$. $\mathbf{A}_s$ is the attention matrix associated with $s$ from the forward pass of LM without fine-tuning. The search gets started by adding the head $h$ as the initial candidate in the beam (line 1). While there are still new candidates waiting to be yielded (line 2), the search continues, and the top $k$ candidates sorted by the matching degrees are maintained (line 3-11) in the beam. In practice, we implement an action manager $\mathcal{O}$ to decide which action to take at each step. Given a candidate $c$ in the beam, $\mathcal{O}(c) = \texttu{start}$ always happens at the beginning of the search. If $c$ has not reached the tail $t$ yet, $\mathcal{O}(c) = \mathbf{\texttu{yield}}$. Otherwise, $\mathcal{O}(c) = \texttu{stop}$. We also notice some facts are in reverse order in the sentence, e.g., {``$\cdots$ said Jason Forcier , a vice president at battery maker A123 Systems Inc.''} for facts of relation {``org:top\_members\_employees''}, thus enable bidirectionality by running the algorithm in both directions (left to right and right to left). The beam search is implemented by the breadth-first search, which is efficient as the time complexity is $O(k \cdot d)$, where $d$ is the maximum depth of the search tree.

\begin{algorithm}[tb]
\small
\vspace{-0.05in}
\caption{\label{alg:beamsearch}{Beam search for matching candidate facts.}}
\begin{algorithmic}[1]
\Require{Head-tail pair $(h,t)$, sentence $s$, attention matrix $\mathbf{A_s}$, action manager $\mathcal{O}=\{\texttu{start}, \texttu{yield}, \texttu{stop}\}$, beam size $k$}
\Ensure{Candidate facts $\mathbb{T}_{(h,t)}$}
\State $\mathbb{T}_{(h,t)} \leftarrow \{\texttu{start}(h)\}$ 
\Comment{{\em Start} by adding the head as a candidate in the beam}
\While{$\exists c \in \mathbb{T}_{(h,t)} [\mathcal{O}(c) = \texttu{yield}]$} 
\State $\widetilde{\mathbb{T}}_{(h,t)} \leftarrow \emptyset$ 
\Comment{Initialize a new beam}
\ForAll{$c \in \mathbb{T}_{(h,t)}$}
\If{$\mathcal{O}(c) = \texttu{yield}$} 
\State $\widetilde{\mathbb{T}}_{(h,t)} \leftarrow  \widetilde{\mathbb{T}}_{(h,t)} \cup \{\texttu{yield}(c,s,\mathbf{A}_s)\}$ \Comment{{\em Yield} a new candidate if not reached the tail}
\Else
\State $\widetilde{\mathbb{T}}_{(h,t)} \leftarrow  \widetilde{\mathbb{T}}_{(h,t)} \cup \{\texttu{stop}(c,t)\}$ \Comment{{\em Stop} then produce a valid fact if reached the tail}
\EndIf
\EndFor
\State $\mathbb{T}_{(h,t)} \leftarrow {\rm TOP}(k, \widetilde{\mathbb{T}}_{(h,t)})$ \Comment{Maintain $k$-best candidates in the beam}
\EndWhile
\State \Return $\mathbb{T}_{(h,t)}$
\end{algorithmic}
\end{algorithm}

\vspace{-0.05in}
\subsubsection{Filter}
\label{sec:filter}
\vspace{-0.05in}

Although the basic functionality provided by beam search is sufficient for finding useful candidate facts, we have found a few constraints useful. Given a candidate fact $(h,r,t)$ from beam search result $\mathbb{T}_{(h,t)}$, it remains as a fact if satisfying all the following constraints.

\vspace{-0.05in}
{\bfseries Constraint \#1 \ } The matching degree of $(h,r,t)$ is above a threshold. We compare the matching degrees corpus-wide to only reserve the facts that are matched better with the knowledge in LMs. For example, \method\ extracts a fact {\sl (Rolling Stone, wrote, pop song)} from {``Rolling Stone wrote: ``No other pop song has so thoroughly challenged artistic conventions''''}, which is not an accurate fact based on the sentence. We observe the associated matching degree is below a proper threshold, while the matching degrees of high-quality facts from the same documents, e.g., {\sl (Dylan, is, songwriter)}, or confident facts from the other documents are beyond the threshold. 

\vspace{-0.05in}
{\bfseries Constraint \#2 \ } The distinct frequency of $r$ is above a threshold. To avoid facts to be over-specified, e.g., {\sl (Dylan, signed to Sam Peckinpah's film, Pat Garrett and Billy the Kid)}, we require $r$ should take many distinct head-tail pairs in the corpus. 

\vspace{-0.05in}
{\bfseries Constraint \#3 \ } Relation $r$ is a contiguous sequence in the sentence. We can avoid $r$ that has no meaningful interpretation~\citep{fader2011identifying}, e.g., {\sl (Rolling Stone, wrote challenged, conventions)} from the above sentence.

\vspace{-0.05in}
\subsection{\steptwo}
\label{sec:kd}

The objective of \steptwo\ stage is to generate an open KG. The open KG contains (a) {\em mapped facts} in a KG schema (Sec.~\ref{sec:ground}), e.g., Wikidata schema, if the schema of the candidate facts is within the existing KG schema; and (b) {\em unmapped facts} from (a) in an open schema (Sec.~\ref{sec:openschema}).

\vspace{-0.05in}
\subsubsection{Mapped Facts in KG Schema}
\label{sec:ground}

The goal is to map a candidate fact $(h, r, t)$ to a fact $(h_k, r_k, t_k)$ in the KG schema. The reason for mapping to an existing KG schema is to make use of the high-quality schema designed by experts (to avoid duplicated efforts of building from scratch) and enable evaluating the candidate facts with oracle KG facts contributed by human volunteers. We first map both entities $h, t$ to $h_k, t_k$, then map the relation $r$ to $r_k$ in the reference KG schema. 

\vspace{-0.05in}
{\bfseries Entity linking to KG schema \ } We leverage an unsupervised entity linker based on a mention-to-entity dictionary~\citep{spitkovsky2012cross} to link the entities for scalability consideration. Besides, contextual information is crucial to link the entities correctly, we use the word embedding of the context to disambiguate the entities, which means we only link the entities with high contextual similarities based on the word embedding. We adopt the entity linker to map $h, t$ to $h_k, t_k$.

\vspace{-0.05in}
{\bfseries Relation mapping with KG schema \ } We largely follow the relation mapping method proposed by \cite{angeli2015leveraging} to construct an offline relation map between KG relation and relation phrases of the candidate facts. The basic idea is that the more often linked head-tail pairs (i.e., entities are with type information from the entity linking step) co-occur between the candidate facts and KG facts, the more likely the corresponding relations are mapped to each other. In addition, we normalize each relation phrase of the candidate facts by lemmatization, and removing inflection, auxiliary verbs, adjectives, adverbs. After the relation mapping is constructed, one author manually checks whether the top 15 relation phrases are true mappings for each KG relation. We only reserve the true ones in the final relation mapping. This process takes approximately one day. Later, $r$ is mapped to $r_k$ with an efficient look-up operation in the relation mapping.

\vspace{-0.05in}
\subsubsection{Unmapped Facts in Open Schema}
\label{sec:openschema}

An unmapped candidate fact $(h, r, t)$ means at least one of $h$, $r$, and $t$ is not mapped to the KG schema based on the method in Sec.~\ref{sec:ground}. There are two types of unmapped candidate facts:

\vspace{-0.05in}
{\bfseries Partially unmapped facts} represent at least one of $h$, $r$, and $t$ are mapped to the KG schema. It can be $h$ or $t$ mapped to $h_k$ or $t_k$ based on the entity linker in Sec.~\ref{sec:ground}. It can also be $r$ that mapped to $r_k$ using the relation mapping in Sec.~\ref{sec:ground}. This actually results in unmapped facts that are in a mixture of the KG schema and the open schema. As the overall schema of the unmapped facts is not in the KG schema, we use open schema to denote such unmapped facts in the rest of the paper for simplicity. An example is {\sl (Dylan, signed, Albert Grossman)} in Figure~\ref{fig:overview}, where both head and tail are linked to Wikidata schema based on the entity linker in Sec.~\ref{sec:ground}, but the relation cannot be mapped since there is no relation mapping from ``signed'' to a KG relation in Wikidata schema. 

\vspace{-0.05in}
{\bfseries Completely unmapped facts} indicate all $h$, $r$, and $t$ are not mapped to the KG schema. This means neither the entity linker nor the relation mapping is able to map $h$, $r$, and $t$ to $h_k$, $r_k$, $t_k$ respectively. The resulting unmapped candidate facts stay in the open schema, e.g., a candidate fact {\sl (Jacob, was, A Registered Mennonite)} stays the same in the open schema from a sentence ``Jacob was a registered Mennonite in Amsterdam.''.

The resulting open KG is a new type of KG that mixes the fixed KG schema with the flexible open schema, suggesting new directions for the next generation of KGs. The open KG not only contains existing knowledge in the reference KGs, but also extends the fixed KG schema with an additional open schema to improve the coverage of the KGs, that benefits the downstream KG based applications, e.g., QA and commonsense reasoning~\citep{wang2019superglue,brown2020language}. 

\vspace{-0.05in}
\section{Experiments}
\label{sec:exp}
\vspace{-0.05in}

\begin{figure*}
    \centering
    \includegraphics[width=0.95\textwidth]{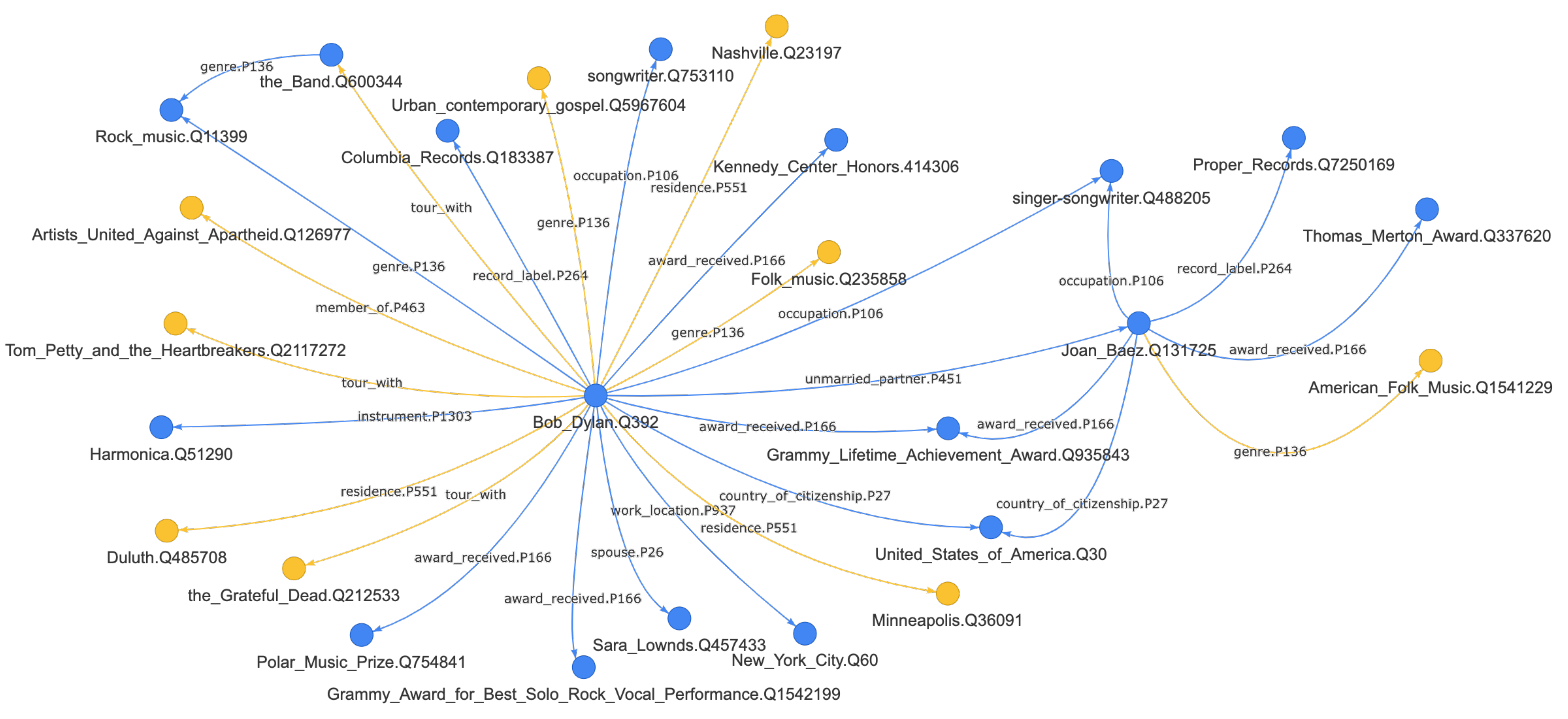}
    \vspace{-0.05in}
    \caption{{\small A snapshot subgraph of the open KG generated by \method\ using BERT$_{\rm LARGE}$ from Wikipedia pages neighboring ``Bob\_Dylan''. The blue node and arrow represent the mapped facts in the Wikidata schema, while the yellow node and arrow denote the unmapped facts in the open schema. We also visualize the correct facts that are new in Wikidata in yellow. }
      \label{fig:kg}}
\end{figure*}

{\it How well can language models generate knowledge graphs?} We experimentally explore how well can \method\ answer the question in the section. To measure the ability of LMs in generating KGs, we directly measure the quality of resulting open KGs. The open KG (see an example in Figure~\ref{fig:kg}) contains two types of facts: mapped facts in the fixed KG schema; and unmapped facts in the open schema. We first quantitatively evaluate \method\ by comparing the mapped facts to oracle KGs annotated by humans in Sec.~\ref{sec:mapped}, then conduct an in-depth analysis of the unmapped facts in Sec.~\ref{sec:unmapped}.

\vspace{-0.05in}
\subsection{Results on Mapped Facts}
\label{sec:mapped}
\vspace{-0.05in}

We first study the quality of the mapped facts. As the candidate facts have been mapped to the schema of oracle KGs, we are able to quantitively compare the candidate facts with the oracle facts in the reference KGs.

\vspace{-0.1in}
\subsubsection{Datasets}
\vspace{-0.05in}

We compare the mapped facts from \method\ with the facts in two KGs:

\vspace{-0.05in}
{\bfseries TAC KBP \ } TAC Knowledge Base Population (KBP) Slot Filling is a task to search a document collection to fill in the tail/object entity for predefined relations (slots) for a given head/subject entity in a reference KG. We experiment with the reference KG in the 2013 challenge. We use the document collection and oracle facts of the 2013 task. The statistic of the dataset is shown in Table~\ref{tab:data}.

\vspace{-0.05in}
{\bfseries Wikidata \ } We use popular Wikidata as another KG. We use all the oracle facts in Wikidata. We use the English Wikipedia as the text corpus, since a large amount of facts in Wikidata is from English Wikipedia. The statistic is in Table~\ref{tab:data}.

\begin{table}[t]
\centering
\resizebox{0.4\linewidth}{!}
  {
  \begin{tabular} {c | c | c }
        \toprule
    {\bf KG} & {\bf \# of oracle facts} & {\bf \# of documents} \\
    \hline 
    TAC KBP                              & 27,655 \footnoteref{ft:anydoc}                             & 3,877,207  
    \\
    Wikidata                      & 27,368,562                    & 6,047,494            \\
    \bottomrule
  \end{tabular}
  }
  \vspace{-0.05in}
\caption{{\small Dataset statistics of two knowledge graphs: TAC KBP and Wikidata. TAC KBP refers to TAC KBP Slot Filling 2013 challenge. \# of oracle facts for TAC KBP is the number of oracle facts in the 2013 task. \# of documents for TAC KBP is the number of the documents in the 2013 task. \# of oracle facts for Wikidata is the total number of oracle facts in Wikidata. \# of documents for Wikidata is the size of English Wikipedia.}}
\label{tab:data}
\end{table}

To evaluate the mapped facts, we first use \stepone\ stage of \method\ to run over the corresponding documents to generate the candidate facts. Then \steptwo\ stage is leveraged to map the candidate facts to the schema of TAC KBP and Wikidata respectively. The parameter settings, such as beam size in Algorithm~\ref{alg:beamsearch}, are shared across TAC KBP and Wikidata based on the parameter study in Sec.~\ref{sec:para}. 

\vspace{-0.05in}
\subsubsection{TAC KBP}
\vspace{-0.05in}

\begin{table}[t]
\centering
\resizebox{0.65\linewidth}{!}
  {
  \begin{tabular} {l | c | c | c}
    \toprule
    {\bf Method} & {\bf Precision\%} & {\bf Recall\%} & {\bf F1\%} \\
    \hline 
OpenIE 5.1 \footnoteref{ft:openie51}                            & 56.98                             & 14.54                              & 23.16                                                \\
Stanford OpenIE \citep{angeli2015leveraging}                             & 61.55                             & 17.35                              & 27.07                                                \\ 
 \hdashline
\method-BERT$_{\rm BASE}$     (\electricblue{\small ours})                                  & 61.57                             & 18.79                              &  28.79                                               \\ 
\method-BERT$_{\rm LARGE}$ (\electricblue{\small ours})                        & 61.69                   & 18.99                     & 29.05                  \\ 
\method-GPT-2 (\electricblue{\small ours})                                    &      61.62                        & 18.17                             & 28.07                                              \\ 
\method-GPT-2$_{\rm MEDIUM}$ (\electricblue{\small ours})                                    & 62.10                             & 18.65                             &   28.69                                             \\
\method-GPT-2$_{\rm LARGE}$ (\electricblue{\small ours})                                    & 62.38                            & 19.00                             &  \secbest 29.12                                              \\
\method-GPT-2$_{\rm XL}$ (\electricblue{\small ours})                                    & 62.69                            & 19.47                             &   \best 29.72                                              \\
    \bottomrule
  \end{tabular}
  }
  \vspace{-0.05in}
\caption{{\small Compare the quality of mapped facts on TAC KBP.}}
\label{tab:resulttac}
\end{table}

To verify the ability to produce correct facts, we compare candidate facts from \method\ to the outputs of two open information systems, which also produce triplets in the form of $(h, r, t)$.

\vspace{-0.05in}
{\bfseries Stanford OpenIE} leverages POS tag and dependency parser, and generates self-contained clauses from long
sentences to extract the triplets, which is the best open information extraction system~\citep{angeli2015leveraging} on TAC KBP~\citep{surdeanu2013overview}. After collecting the triplets from the system, we use the same \steptwo\ procedure with \method\ (Sec.~\ref{sec:ground}) to map the triplets to the corresponding KG schema.

\vspace{-0.05in}
{\bfseries OpenIE 5.1}~\footnote{\label{ft:openie51}\tiny\url{https://github.com/dair-iitd/OpenIE-standalone}} is the successor to Ollie~\citep{schmitz2012open}, and it improves extractions from noun relations, numerical sentences, and conjunctive sentences depending on the linguistic patterns. Similarly, the same \steptwo\ procedure (Sec.~\ref{sec:ground}) is leveraged to map the triplets from the system to the KG schema.

We use two families of pre-trained LMs with \method.
We use BERT$_{\rm BASE}$ and BERT$_{\rm LARGE}$ from \cite{devlin2018bert} with \method, namely {\bf \method-BERT$_{\rm BASE}$} and {\bf \method-BERT$_{\rm LARGE}$}. 
Besides, GPT-2s from \cite{radford2019language} are used, i.e., {\bf \method-GPT-2}, {\bf \method-GPT-2$_{\rm MEDIUM}$}, {\bf \method-GPT-2$_{\rm LARGE}$}, and {\bf \method-GPT-2$_{\rm XL}$}.

Table~\ref{tab:resulttac} shows the results on TAC KBP. We use the official scorer of TAC KBP Slot Filling 2013 to evaluate precision, recall, and F1 on TAC KBP~\footnote{\label{ft:anydoc}\tiny{There are 2,383 correct oracle facts based on the ``manual runs'' assessment in TAC KBP.}}. 

\vspace{-0.05in}
{\bfseries \method\ constructs improved KGs}. From the results, we find that all our methods achieve competitive precision, which is greater than 60\%, given the unsupervised nature of \method. All the proposed methods outperform the two open IE systems. This shows that \method\ is able to produce high-quality knowledge directly from pre-trained LMs by a single forward pass without human supervision. The results show the effectiveness of \method\ in generating candidate facts from \stepone\ stage, and producing high-quality KGs through \steptwo\ stage. We also find that \method-GPT-2$_{\rm XL}$ performs the best. \method-GPT-2$_{\rm XL}$ outperforms the previous state-of-the-art Stanford OpenIE by over 2.6\% in F1. This shows the proposed end-to-end \method\ is able to recover the knowledge stored in pre-trained LMs without relying on any extra linguistic features, such as POS tag and dependency parser used in open IE systems. The main reason leading to the moderate results of OpenIE 5.1 is that the system generates objects of the triplets with extraneous words, which hurt the performance in slot filling tasks. Even though the proposed methods all outperform the two open IE systems in the recall, however improving recall is clearly the future direction to further improve the performance of \method. We find that the main cause of the moderate recall is the incorrect entities caused by spaCy noun chunk as summarized in Sec.~\ref{sec:error}.

\vspace{-0.05in}
{\bfseries Larger/deeper LMs produce KGs of higher quality}. BERT$_{\rm LARGE}$ outperforms BERT$_{\rm BASE}$ since the doubling parameter size. GPT-2s share similar trends, where we observe performance increases when the model size increases. This complies with our intuition on more knowledge is stored in deeper and larger models. Such increases in performance seem subtle on TAC KBP, we find this might due to the relatively small number of oracle facts by noticing a more significant improvement on Wikidata in Sec.~\ref{sec:wiki}. We plan to further improve the results with larger pre-trained LMs, e.g., GPT-3~\citep{brown2020language}, Megatron-LM~\citep{shoeybi2019megatron}. 

\vspace{-0.05in}
{\bfseries BERT LMs outperform GPT-2 LMs under similar model sizes}. More specifically, BERT$_{\rm BASE}$ performs better than \method-GPT-2 in F1, and \method-BERT$_{\rm LARGE}$ outperforms \method-GPT-2$_{\rm MEDIUM}$ in F1. BERT$_{\rm BASE}$ and \method-GPT-2 are similar in size, while \method-BERT$_{\rm LARGE}$ and \method-GPT-2$_{\rm MEDIUM}$ are similar in model size as well. This is mainly because that the recall of BERT LMs is higher than that of corresponding GPT-2 LMs. The result indicates that the Cloze-style loss function (i.e., masked language model) of BERT is more effective and flexible in recovering more knowledge than the autoregressive LM objective. We also notice that the precision of GPT-2 LMs is higher than that of according BERT LMs. The reason is that the autoregressive LM objective captures more accurate knowledge than Cloze-style loss does by not introducing extra noise (e.g., masked tokens) during pre-training.

\vspace{-0.05in}
\subsubsection{Wikidata}
\label{sec:wiki}
\vspace{-0.05in}

We select our best BERT based method \method-BERT$_{\rm LARGE}$, and GPT-2 based method \method-GPT-2$_{\rm XL}$ on TAC KBP to compare with Stanford OpenIE (the best open IE system on TAC KBP) for scalability experiments on Wikidata. We follow the same definition as the slot filling task to calculate precision, recall, and F1 on Wikidata. Table~\ref{tab:resultwiki} summarizes the results. 

\vspace{-0.05in}
{\bfseries \method\ is scalable to larger corpora}. Similar to the trends on TAC KBP, \method-GPT-2$_{\rm XL}$ performs the best in precision, recall, and F1. The results show the effectiveness of \method\ in generating candidate facts and high-quality KGs. We also find that \method-GPT-2$_{\rm XL}$ outperforms \method-BERT$_{\rm LARGE}$ by over 1\% in F1. This shows that the larger model (GPT-2$_{\rm XL}$ has 5x more parameters compared to BERT$_{\rm LARGE}$) contains more knowledge, and \method\ is able to restore the knowledge. When larger or deeper models (e.g., GPT-3) are used with \method, we can expect more gains of the KG quality. Thanks to the efficient nature of \method, which relies only on the forward pass of the LMs without fine-tuning, the results suggest that \method\ is scalable to large KGs. 

\vspace{-0.05in}
{\bfseries Larger corpora embed more complete KGs}. In particular, \method-GPT-2$_{\rm XL}$ outperforms Stanford OpenIE by 5.6\% in F1. \method-BERT$_{\rm LARGE}$ outperforms Stanford OpenIE by approximately 4.4\% in F1. Both F1 gains are larger compared to that on TAC KBP. This is because that the LMs contain world knowledge from pre-training corpora, e.g. Wikipedia and Common Crawl. The larger the textual corpora are, the more knowledge our method is able to recover and match to the knowledge stored in LMs. The finding is particularly important, since we are now able to construct larger KGs of high quality from scratch when larger datasets are used, such as WebText2 and Common Crawl~\citep{abs-1910-10683,brown2020language}. Similar to the observations on TAC KBP, the precision is higher compared to recall. Wikidata is not fully built from Wikipedia, \method\ could improve the recall by running on those larger corpora to collect more facts.

\begin{table}[t]
\centering
\resizebox{0.65\linewidth}{!}
  {
  \begin{tabular} {l | c | c | c }
    \toprule
    {\bf Method} & {\bf Precision\%} & {\bf Recall\%} & {\bf F1\%} \\
    \hline 
    Stanford OpenIE \citep{angeli2015leveraging}                              & 23.32                             & 13.09                              & 16.77 
    \\
    \hdashline
    \method-BERT$_{\rm LARGE}$ (\electricblue{\small ours})                       & 29.52                    & 16.56                    & \secbest 21.22                 \\
    \method-GPT-2$_{\rm XL}$ (\electricblue{\small ours})                       & 31.32                    & 17.42                    & \best 22.39                \\
    \bottomrule
  \end{tabular}
  }
  \vspace{-0.05in}
\caption{{\small Compare the quality of mapped facts on Wikidata.}}
\label{tab:resultwiki}
\end{table}

\vspace{-0.05in}
\subsection{Analysis of Unmapped Facts}
\label{sec:unmapped}
\vspace{-0.05in}

As illustrated in Figure~\ref{fig:kg}, the open KG constructed by \method\ is a new type of KG combining the fixed KG schema with the flexible open schema. We turn to study the quality of the candidate facts that are not mapped to the above reference KG schema, but are in the open schema generated by \method. We manually judge such unmapped facts generated by our best method \method-GPT-2$_{\rm XL}$ from 100 sampled documents in Wikidata and TAC KBP respectively.

\vspace{-0.05in}
{\bfseries The quality of unmapped facts is verified}. We find 35.3\% of the unmapped facts are true on Wikidata. We find 83.2\% of those true facts are partially unmapped facts as defined in Sec.~\ref{sec:openschema}, e.g., {\sl (Bob\_Dylan.Q392, tour\_with, the\_Grateful\_Dead.Q212533)}, whose relation is not within the schema of Wikidata, while both head and tail are in the schema. The remaining true facts are completely unmapped facts (Sec.~\ref{sec:openschema}), e.g., a candidate fact {\sl (Jacob, was, A Registered Mennonite)} stays the same in the open schema. 

\vspace{-0.05in}
{\bfseries Accurate entity detection is desired}. We also notice 45.5\% of the untrue unmapped facts on Wikidata are due to the {\em incorrect entities} detected by the spaCy. {\em Incorrect or missing entity linking} (to either head or tail) in Sec.~\ref{sec:ground} causes additional {9.1\%} errors in the unmapped facts. {4.5\%} of the untrue unmapped facts are caused by the {\em missing relation mapping} in Sec.~\ref{sec:ground}. The rest errors made by \method-GPT-2$_{\rm XL}$ are {\em incorrect relation phrases}, such as uninformative relation phrases, e.g., {\sl (Dylan, made, his breakthrough)}, which is similar to the errors made by open IE systems~\citep{fader2011identifying}. Both entity linking and relation mapping of \steptwo\ stage rely heavily on the accuracy of entity detection from the spaCy noun chunk. We conclude that the main root cause of the untrue unmapped facts is due to the errors made by the spaCy noun chunk.

We observe similar trends on TAC KBP. We plan to leverage crowdsourcing platforms, e.g., Mechanical Turk, to conduct quantitative evaluations over the unmapped facts to better understand the strengths and shortage of \method. We plan to identify more accurate entities by relying on attention weights in LMs~\citep{clark2019does,hewitt2019structural} instead of using extra resources. We will also investigate stronger entity linkers~\citep{kolitsas2018end} and learn a more robust relation mapping through weak or distant supervision~\citep{mintz2009distant,ratner2017snorkel}. We will investigate more sophisticated approaches, such as graph neural networks~\citep{kipf2016semi}, to generate more accurate relation phrases from the attention weight matrices by considering structural information.

\vspace{-0.2in}
\section{Related Work}
\vspace{-0.15in}

{\bf Knowledge graph construction} can be generally categorized into two groups, 1) supervised approaches. Wikidata, Freebase~\citep{bollacker2008freebase}, YAGO~\citep{suchanek2007yago}, YAGO2~\citep{hoffart2013yago2}, DBpedia~\citep{auer2007dbpedia} are built based on human supervision from Wikipedia infoboxes and other structured data sources; 2) semi-supervised approaches. Open information extraction systems, e.g., OLLIE~\citep{schmitz2012open}, Reverb~\citep{fader2011identifying}, Stanford OpenIE~\citep{angeli2015leveraging}, and OpenIE 5.1 \footnoteref{ft:openie51} aim to leverage carefully-designed patterns based on linguistic features (e.g., dependencies and POS tags), to extract triplets from web corpora for open schema KG. Besides, NELL~\citep{carlson2010toward}, DeepDive~\citep{niu2012elementary}, Knowledge Vault~\citep{dong2014knowledge} extract information based on a fixed schema or ontology, where humans help improve the accuracy of the extractions. Probase~\citep{wu2012probase} produces taxonomies instead of rich typed relations in general KGs. \method\ instead uses learned knowledge stored in pre-trained LMs without human supervision to construct an open KG, which is a mixture of fixed schema and open schema.

\vspace{-0.05in}
{\bf Language models}, e.g., BERT~\citep{devlin2018bert}, GPT~\citep{radford2018improving}, GPT-2/3~\citep{radford2019language,brown2020language}, ELMo~\citep{peters2018deep}, Transformer-XL~\citep{dai2019transformer}, ALBERT~\citep{abs-1909-11942}, RoBERTa~\citep{abs-1907-11692}, XLNet~\citep{abs-1906-08237} and Megatron-LM~\citep{shoeybi2019megatron} contain factual knowledge obtained via pre-training on large-scale corpora such as Wikipedia and BookCorpus~\citep{zhu2015aligning}. Studies have leveraged the pre-trained LMs as virtual KGs, and show reasonable performance in QA tasks~\citep{dhingra2020differentiable,guu2020realm}, and language modeling~\citep{khandelwal2019generalization}. LMs are further enhanced by KGs~\citep{zhang2019ernie,peters2019knowledge} to improve knowledge-driven tasks. While the existing work utilizes knowledge in an implicit way, the main difference is that our approach explicitly extracts knowledgeable facts from the LMs. Compare to LAMA~\citep{petroni2019language,petroni2020context}, instead of conducting fact recall in cloze statements, \method\ directly generates the whole fact in the form of a triplet $(h, r, t)$ from a sentence. Besides, the benchmark datasets used with \method\ are larger compared to the LAMA benchmark, e.g., Wikidata is 3 orders of magnitude larger compared to the largest dataset in the LAMA benchmark.

\vspace{-0.05in}
{\bf Neural network interpretation} here specifically refers to pre-trained deep language model analysis. There has been a lot of work to understand what the neural networks learn~\citep{linzen2016assessing,adi2016fine,tenney2019you}. With regards to analyzing Transformer~\citep{vaswani2017attention} based language models (e.g., BERT and GPT-3), substantial recent work focuses on both visualizing and analyzing the attention~\citep{vig2019visualizing,jain2019attention,clark2019does,michel2019sixteen,vig2020bertology,ramsauer2020hopfield,hendrycks2020measuring}. Instead of analyzing or visualizing, we use LMs to generate structured KGs to directly recover what LMs learn from the corpora.

\vspace{-0.15in}
\section{Conclusion}
\vspace{-0.15in}
We show that the knowledge graphs can be constructed by a single forward pass of the language models over textual corpora. We propose a two-stage unsupervised approach \method\ to first match the facts in the corpus with the internal knowledge of the language model, and then map the matched facts to produce a knowledge graph. We demonstrate the quality of the resultant open knowledge graphs by comparing to two knowledge graphs (Wikidata and TAC KBP). The open knowledge graph also features new facts in the open schema, which could have broad implications for knowledge graphs and their downstream applications. The results also suggest that larger language models store richer knowledge than existing knowledge graphs, and generating on even larger high-quality text corpora could continue improving knowledge graphs. Additionally, the knowledge graphs generated by our approach can help researchers to look into what the language models learn, so our interpretable knowledge graphs establish a bridge between the deep learning and knowledge graph communities.

\section*{Acknowledgements} 
We would like to thank Xinyun Chen, Yu Gai, Dan Hendrycks, Qingsong Lv, Yangqiu Song, Jie Tang, and Eric Wallace for their helpful inputs. We also thank Gabor Angeli, Christopher D. Manning for their timely help in replicating the results in \cite{angeli2015leveraging}. This material is in part based upon work supported by Berkeley DeepDrive and Berkeley Artificial Intelligence Research.

\bibliography{iclr2021_conference}
\bibliographystyle{iclr2021_conference}

\newpage 
 
\appendix
\section{Additional Details and Analysis of \method}

\subsection{Implementation Details}
\label{sec:implement}

To evaluate the mapped facts, we first use \stepone\ stage of \method\ to run over the corresponding documents to generate the candidate facts. For \steptwo\ stage on TAC KBP, we link to the oracle annotation of the entities or spans in the TAC KBP corpus. On Wikidata, the entity linking method described in Sec.~\ref{sec:ground} is first leveraged to link entities in the candidate facts to Wikipedia anchors. Then a Wikipedia anchor to the Wikidata item dictionary is used to further link the entities to Wikidata. If the head or tail is a pronoun, we further use neuralcoref~\footnote{\tiny\url{https://github.com/huggingface/neuralcoref}} for coreference resolution. We use GloVe~\citep{pennington2014glove} embedding for disambiguation. The relation mapping is constructed offline for TAC KBP and Wikidata respectively using the method in Sec.~\ref{sec:ground}. For oracle facts in Wikidata, we only preserve those facts describing relations that could be linked to a corresponding Wikipedia anchor. We rule out facts of attributes about entities and facts of auxiliary relations (such as \textit{topic's main category.P901}) and finally results in 27,368,562 oracle facts.

For Wikidata, at \stepone\ stage, we randomly split the English Wikipedia data into 20 partitions, and map the data partitions to 20 distributed servers to run. Each server is configured with four Tesla K80 12Gs. We set the max sequence length to 256, and batch size as 32 for \method-BERT$_{\rm LARGE}$ and 4 for \method-GPT-2$_{\rm XL}$. We use implementations of pre-trained LMs in Transformers package~\footnote{\tiny\url{https://github.com/huggingface/transformers}}. We use spaCy sentencizer~\footnote{{\tiny\url{https://spacy.io/api/sentencizer}}} to segment the documents into sentences. \method-BERT$_{\rm LARGE}$ takes approximately 48 hours, and \method-GPT-2$_{\rm XL}$ costs around 96 hours. The resulting candidate facts of \stepone\ stage from the 20 servers are then reduced a data server, where a MongoDB database is maintained to store the oracle Wikidata and entity linking results to enable the efficient \steptwo\ stage. To produce the open KGs, \steptwo\ stage takes around 18 hours. The setup is similar to TAC KBP. \stepone\ stage is done within 48 hours for all the settings. The batch sizes of \method-BERT$_{\rm BASE}$, \method-GPT-2, \method-GPT-2$_{\rm MEDIUM}$, \method-GPT-2$_{\rm LARGE}$ are 64, 32, 16, 8 respectively.

The parameter settings are shared across TAC KBP and Wikidata. All the choices are based on the parameter study in Sec.~\ref{sec:para}. The beam size of Algorithm~\ref{alg:beamsearch} is set to 6. The matching degree threshold of Constraint \#1 (Sec.~\ref{sec:filter}) is set to 0.005, and the number of distinct head-tail pairs of Constraint \#2 (Sec.~\ref{sec:filter}) is set to 10. To generate the attention weight matrix $\mathbf{A}_s$ of a sentence, we reduce the weights of every attention head in the last layer of pre-trained LMs using the mean operator.  

\subsection{Error Analysis}
\label{sec:error}
There is still significant room to improve \method. To further understand the shortage of \method, we conduct an error analysis of the errors in precision (i.e., incorrect facts returned by \method) of Table~\ref{tab:resulttac} and Table~\ref{tab:resultwiki}. We choose our best method \method-GPT-2$_{\rm XL}$ for the study. We sample 100 documents from the Wikidata dataset, and manually check the reasons for the errors. We find {33.1\%} of the errors are caused by {\em incorrect entities}, while the relation phrases are correct. The errors are due to the incorrect noun chunk detected by the spaCy~\footnote{\tiny{\url{https://spacy.io/usage/linguistic-features/\#noun-chunks}}}. {18.3\%} of the errors are due to the {\em missing relation mapping} created in Sec.~\ref{sec:ground}. Note that we find approximately {23.8\%} of the errors are actually {\em correct facts that are new in the reference KGs}. e.g., {\sl (Bob\_Dylan.Q392, residence.P551, Nashville.Q23197)} (in Figure~\ref{fig:kg}) is not an existing fact in Wikidata, but it is a correct mapped fact based on our annotation. The rest errors made by \method-GPT-2$_{\rm XL}$ are {\em incorrect relation phrases}, such as uninformative relation phrases. We find similar errors are made by \method-GPT-2$_{\rm XL}$ on TAC KBP. 
Similar to Sec.~\ref{sec:unmapped}, enhancing the entity detection, entity linker, relation mapping, and relation generation are helpful.
We also plan to leverage lifelong learning~\citep{carlson2010toward} to add true facts to the reference KGs to improve the evaluation. 

\subsection{Parameter Study}
\label{sec:para}

\begin{figure*}
\centering
\subcaptionbox{{\small Beam size.}}{\includegraphics[width=0.3\textwidth]{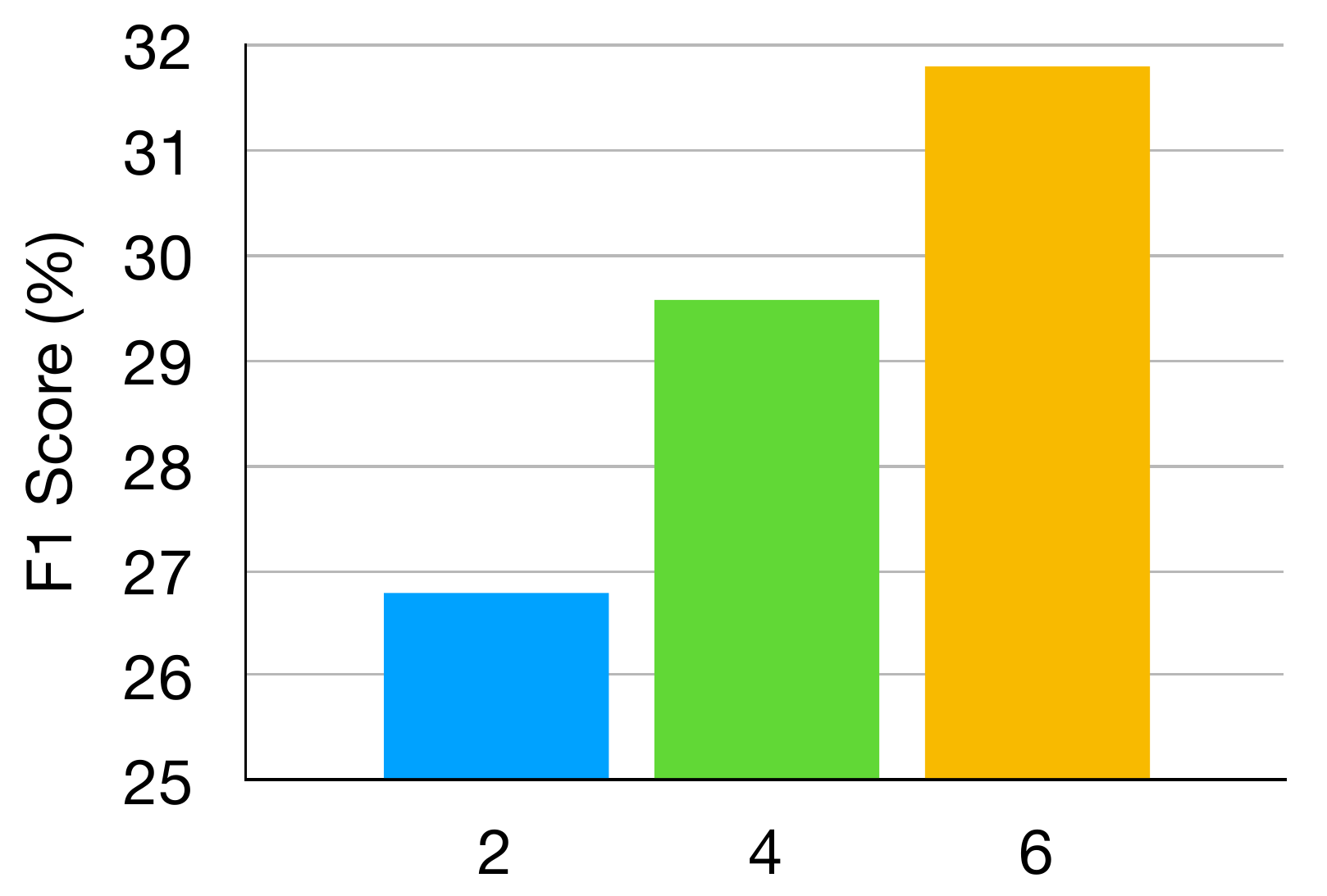}}%
\hspace{0.1in}
\subcaptionbox{{\small Constraint \#1.}}{\includegraphics[width=0.3\textwidth]{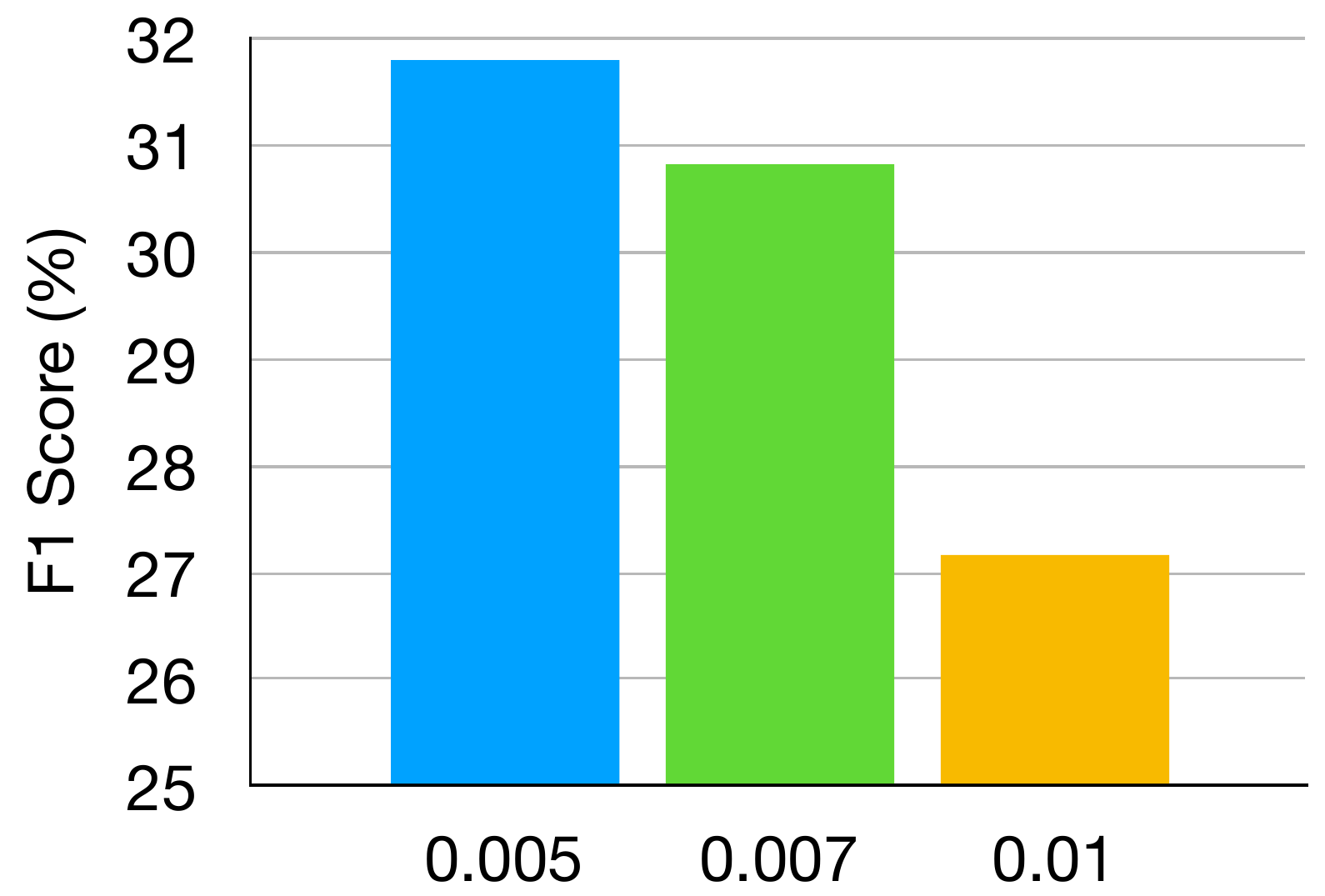}}%
\hspace{0.1in}
\subcaptionbox{{\small Constraint \#2.}}{\includegraphics[width=0.3\textwidth]{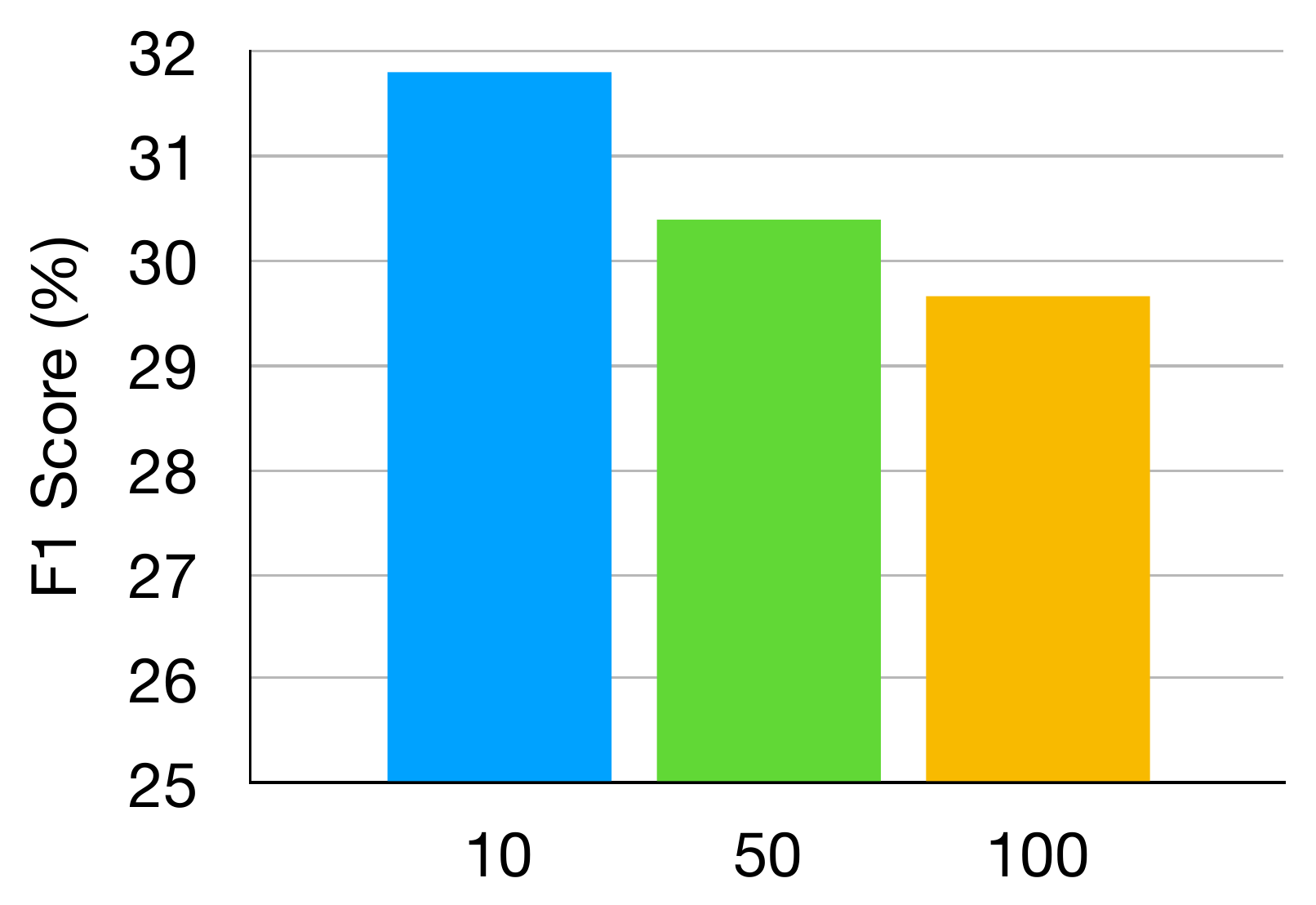}}%
\\
\subcaptionbox{{\small Attention weights layers.}}{\includegraphics[width=0.32\textwidth]{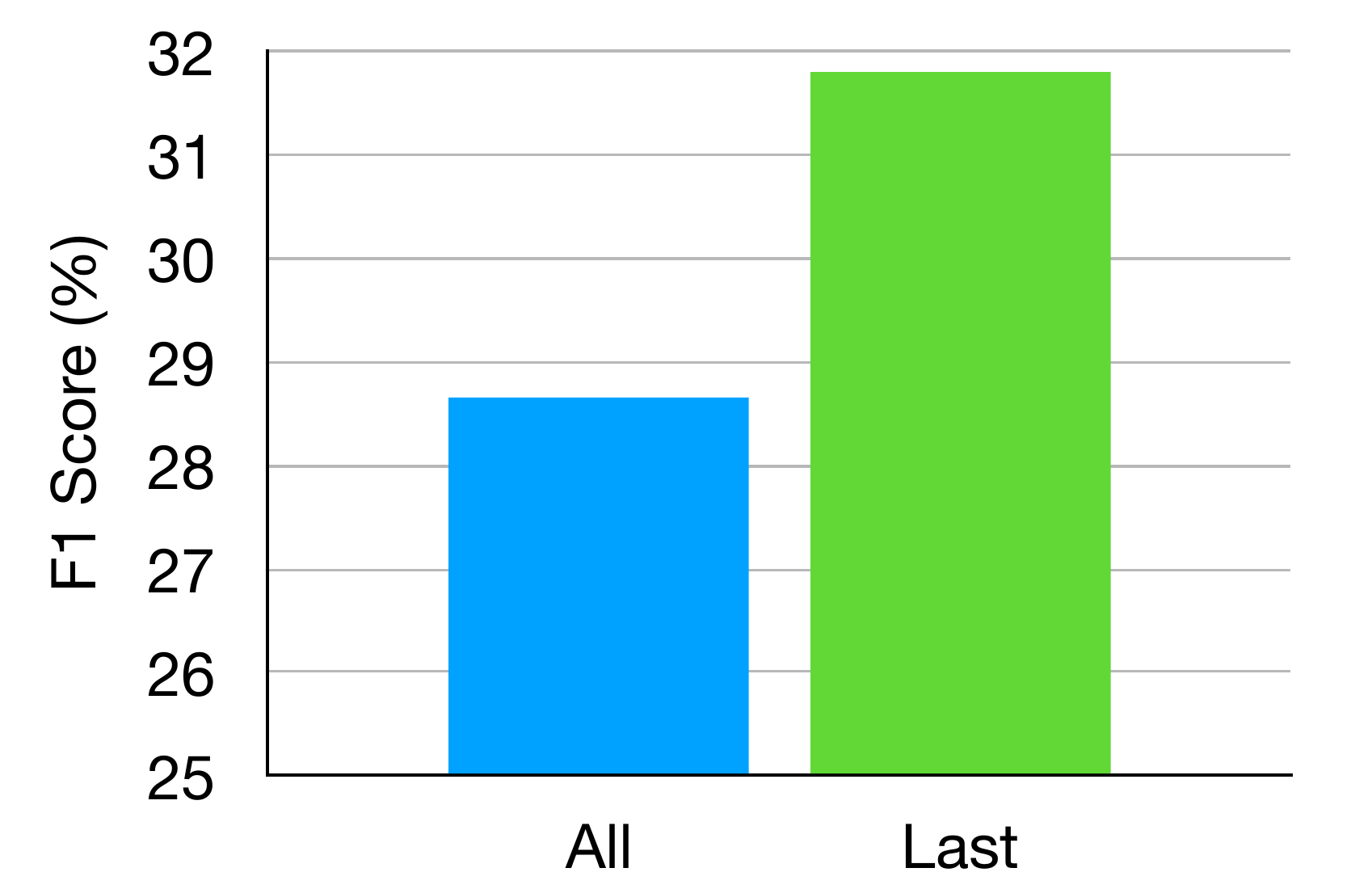}}%
\hspace{0.1in}
\subcaptionbox{{\small Attention weights reduction.}}{\includegraphics[width=0.3\textwidth]{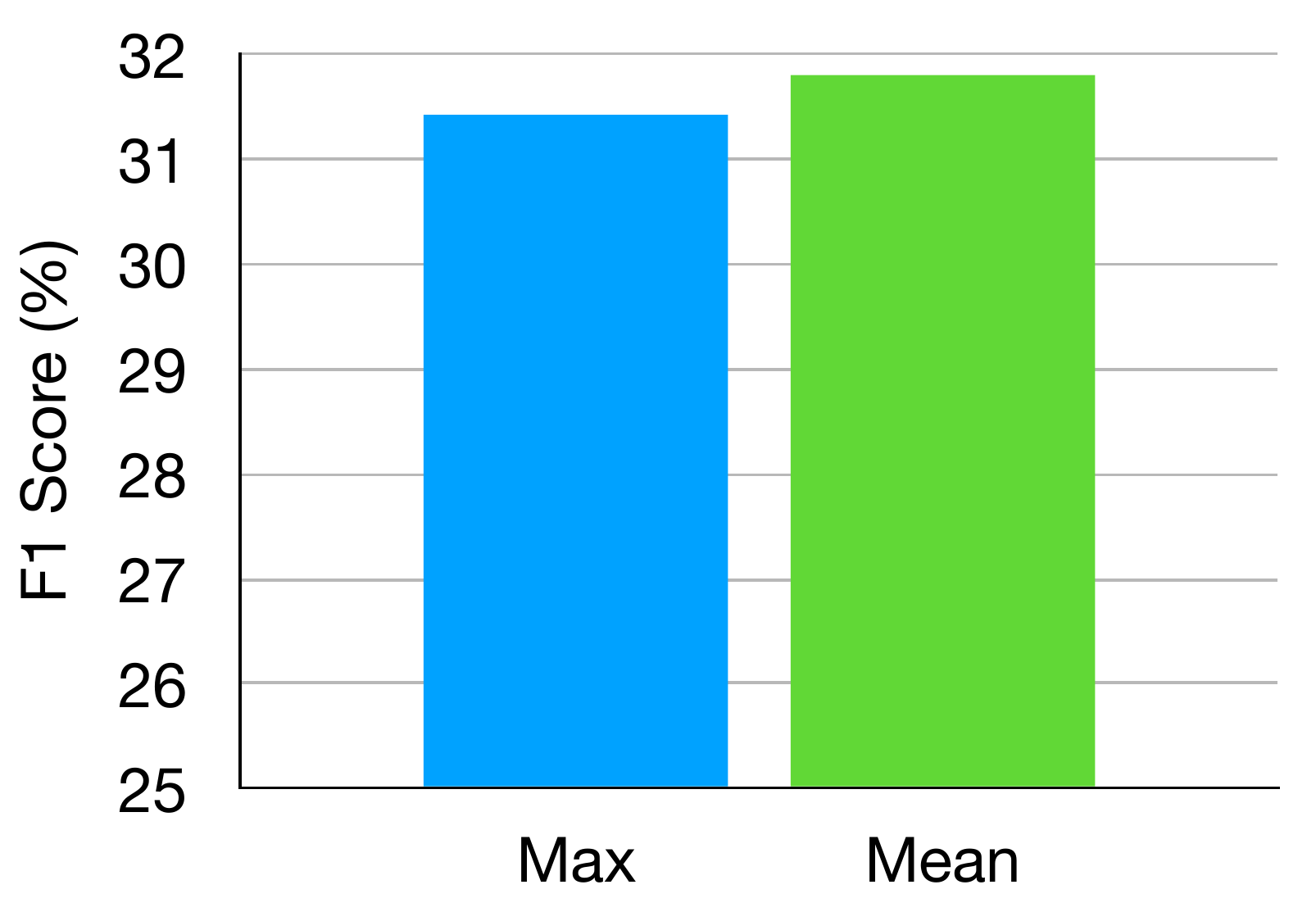}}%
\caption{{\small Parameter study with \method-BERT$_{\rm BASE}$ on TAC KBP hold-out subset.}}
\label{fig:para}
\end{figure*}

We study the effects of the parameters using \method-BERT$_{\rm BASE}$ on TAC KBP. We randomly sample 20\% of the oracle query entities as a hold-out dataset to tune the parameters, and use the best parameter setting achieved for both TAC KBP and Wikidata experiments. When studying the effect of a certain parameter, we keep the remaining parameters as default described in Sec.~\ref{sec:implement}. We use F1 to measure the effects.

{\bfseries Effects of beam size \ } Figure~\ref{fig:para}(a) illustrates the effects of various beam sizes in Algorithm~\ref{alg:beamsearch}. We find that in general, the larger the beam size is, the better F1 the setting achieves. This is because that \method\ is able to reserve more potentially correct facts when more candidates are allowed in the \stepone\ stage. However, F1 improvement gradually becomes subtle, while the computation costs increase more significantly. For sake of the efficiency, we do not explore larger beam sizes. We set the beam size as 6.

{\bfseries Effects of search constraints \ } Figure~\ref{fig:para}(b) compares the effect of different thresholds of the matching degree of Constraint \#1 in Sec.~\ref{sec:filter}. We set the threshold as 0.005 since it achieves the best result. Note that the summed attention score is normalized by the length of the fact to penalize the cumbersome facts. Figure~\ref{fig:para}(c) shows the impact of the number of distinct head-tail pairs in identifying common relations of Constraint \#2 in Sec.~\ref{sec:filter}. The best result is achieved when it equals 10. This shows that while \method\ mostly identifies frequent relations, it is also able to capture some rare relations for the open schema.

{\bfseries Effects of attention weights \ } Figure~\ref{fig:para}(d) shows the comparison between attention weights of the last layer and the mean of all layers. The attention weights of the last layer perform better. This is due to the attention weights in lower layers are low-level linguistic knowledge according to \citep{clark2019does,ramsauer2020hopfield}, which are less relevant to the factual knowledge for the KG construction. Figure~\ref{fig:para}(e) compares the impact of different attention reduction, i.e., mean, max, over the attention heads of the last layer. We find the ``mean'' perform better. The reason is that the token often intensively attends to several specific tokens in the sequence~\citep{michel2019sixteen}, and the ``mean'' operator is more sensitive to such information.

\section{Samples from \method\ on TAC KBP}

\subsection{Mapped Facts}

We randomly sample 100 documents from TAC KBP corpus, then randomly sample sentences from those documents. 
The uncurated candidate facts and the corresponding mapped facts of the sampled sentences based on our best methods \method-BERT$_{\rm LARGE}$ and \method-GPT-2$_{\rm XL}$ are shown in Figure~\ref{fig:kbpmapbert} and Figure~\ref{fig:kbpmapgpt} respectively.
We also randomly sample several sentences in which \method-BERT$_{\rm LARGE}$ differs from \method-GPT-2$_{\rm XL}$ in the resulting facts for comparison, which is illustrated in Figure~\ref{fig:kbpmapbertgpt}.
In each table, ``{\bf ID}'' represents the document ID of a sampled sentence in TAC KBP corpus. ``{\bf Sentence}'' indicates the sampled sentence. ``{\bf Candidate facts to mapped facts}'' column contains the candidate facts (on the left side of ``$\rightarrow$'') and their corresponding mapped facts (on the right side of ``$\rightarrow$'').

\subsection{Unmapped Facts}
We randomly sample 100 documents from TAC KBP corpus. From those documents, we show unmapped facts from the sampled sentences from those documents. We manually check the correctness of the unmapped facts according to Sec.~\ref{sec:unmapped}, and show the correct ones.
The original candidate facts with the corresponding unmapped facts of the sampled sentences generated by \method-BERT$_{\rm LARGE}$ and \method-GPT-2$_{\rm XL}$ are shown in Figure~\ref{fig:kbpunmapbert} and Figure~\ref{fig:kbpunmapgpt}. A further comparison of the unmapped candidate facts is illustrated in Figure~\ref{fig:kbpunmapbertgpt}. 
In each table, ``{\bf ID}'' represents the document ID of a sampled sentence in TAC KBP corpus. ``{\bf Sentence}'' indicates the sampled sentence. ``{\bf Candidate facts to unmapped facts}'' column contains the candidate facts (on the left side of ``$\rightarrow$'') and their corresponding unmapped facts (on the right side of ``$\rightarrow$'').

\section{Samples from \method\ on Wikidata}
\label{seca:wiki}

\subsection{Mapped Facts}

Similar to TAC KBP, we randomly sample 100 documents from the Wikidata corpus (i.e., English Wikipedia), then randomly sample sentences from those documents. 
The uncurated candidate facts and the corresponding mapped facts of the sampled sentences based on our best methods \method-BERT$_{\rm LARGE}$ and \method-GPT-2$_{\rm XL}$ are shown in Figure~\ref{fig:wikimapbert} and Figure~\ref{fig:wikimapgpt} respectively.
We also randomly sample several sentences in which \method-BERT$_{\rm LARGE}$ differs from \method-GPT-2$_{\rm XL}$ in the resulting facts for comparison, which is illustrated in Figure~\ref{fig:wikimapbertgpt}.
In each table, ``{\bf ID}'' represents the Wikipedia page's title of a sampled sentence. ``{\bf Sentence}'' indicates the sampled sentence. ``{\bf Candidate facts to mapped facts}'' column contains the candidate facts (on the left side of ``$\rightarrow$'') and their corresponding mapped facts (on the right side of ``$\rightarrow$'').

\subsection{Unmapped Facts}

Similar to TAC KBP, we randomly sample 100 documents from the Wikidata corpus. From those documents, we show unmapped facts from several sampled sentences from those documents. We manually check the correctness of the unmapped facts according to Sec.~\ref{sec:unmapped}, and show the correct ones.
The original candidate facts with the corresponding unmapped facts of the sampled sentences generated by \method-BERT$_{\rm LARGE}$ and \method-GPT-2$_{\rm XL}$ are shown in Figure~\ref{fig:wikiunmapbert} and Figure~\ref{fig:wikiunmapgpt}. A further comparison of the unmapped candidate facts is illustrated in Figure~\ref{fig:wikiunmapbertgpt}. 
In each table, ``{\bf ID}'' represents the Wikipedia page's title of a sampled sentence. ``{\bf Sentence}'' indicates the sampled sentence. ``{\bf Candidate facts to unmapped facts}'' column contains the candidate facts (on the left side of ``$\rightarrow$'') and their corresponding unmapped facts (on the right side of ``$\rightarrow$'').

\section{Additional Open KG Subgraphs from \method\ on Wikidata}
\begin{figure*}
    \centering
    \includegraphics[width=0.5\textwidth]{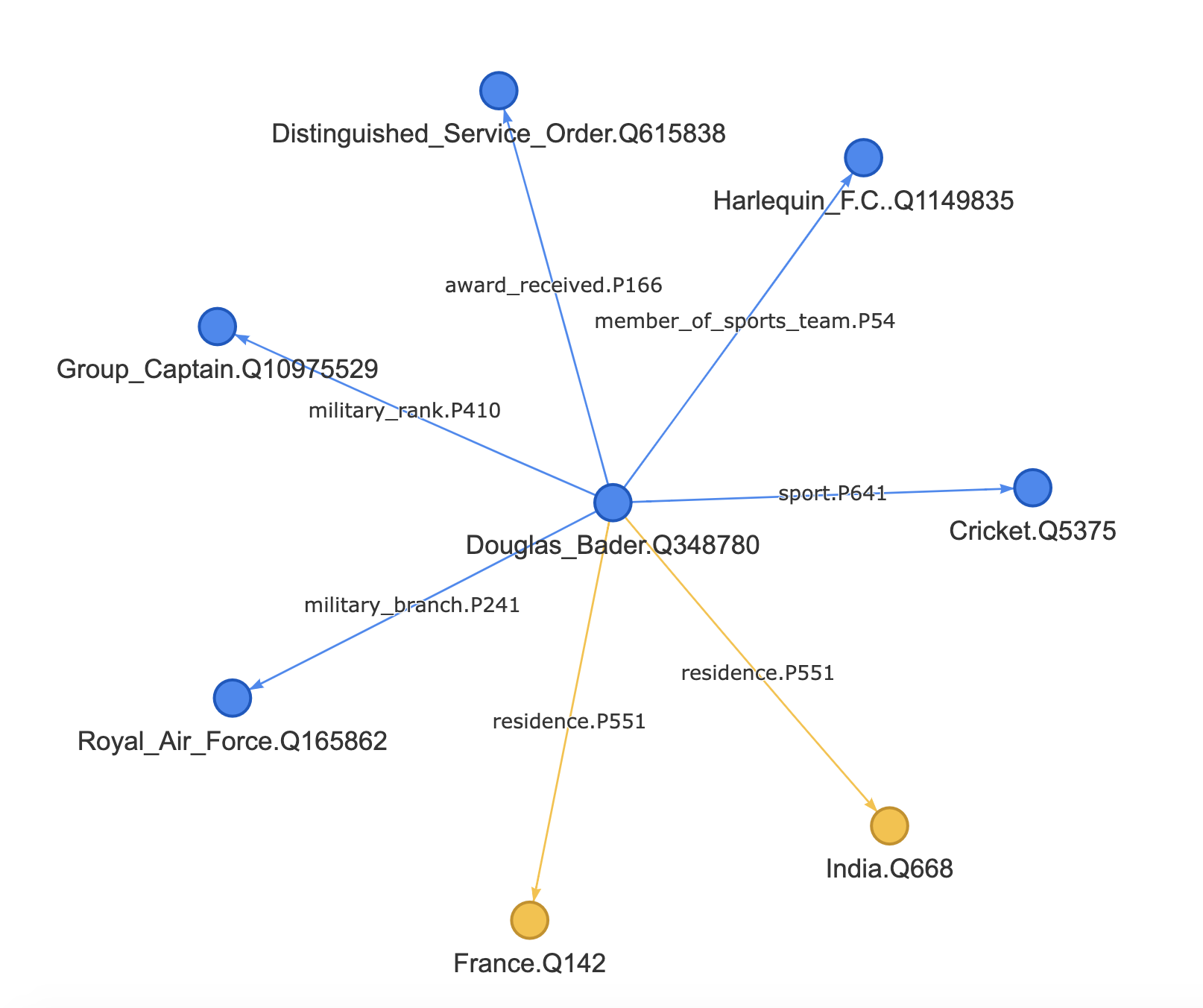}
    \caption{{\small A snapshot subgraph of the open KG generated by \method-BERT$_{\rm LARGE}$ from the Wikipedia page ``Douglas\_Bader''. }
      \label{fig:kg1}}
\end{figure*}

\begin{figure*}
    \centering
    \includegraphics[width=0.4\textwidth]{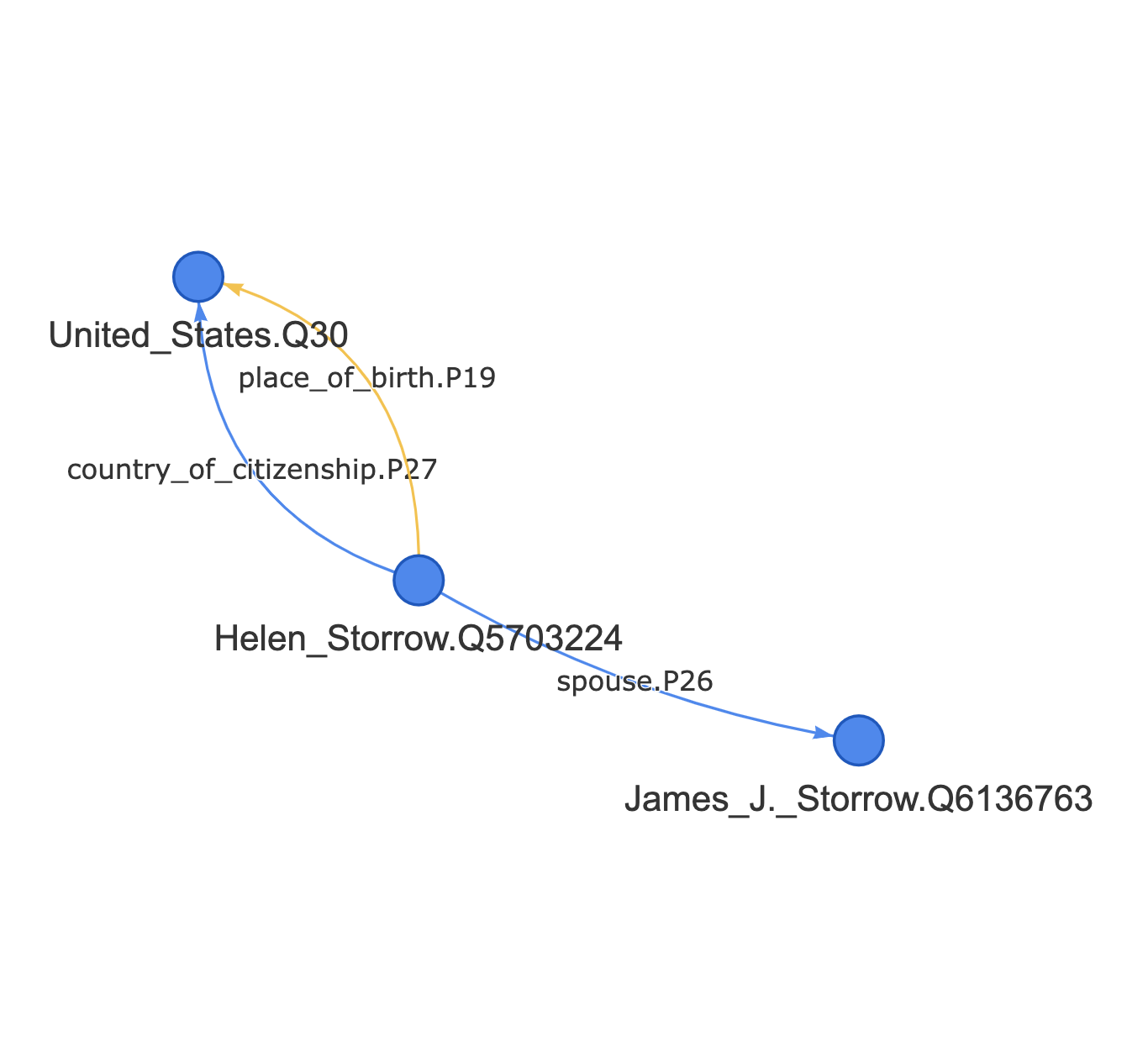}
    \caption{{\small A snapshot subgraph of the open KG generated by \method-BERT$_{\rm LARGE}$ from the Wikipedia page ``Helen\_Storrow''. }
      \label{fig:kg3}}
\end{figure*}

\begin{figure*}
    \centering
    \includegraphics[width=0.5\textwidth]{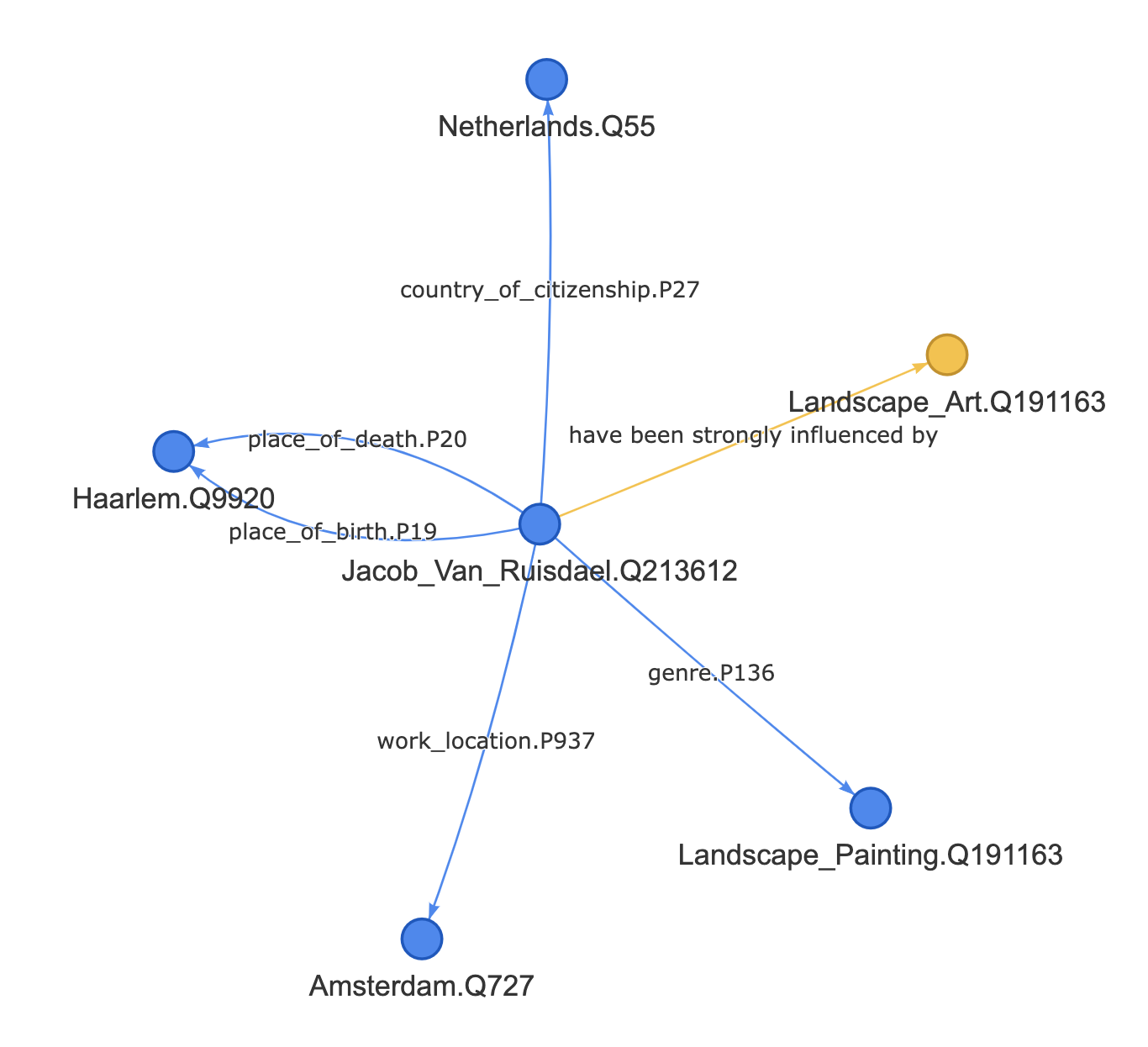}
    \caption{{\small A snapshot subgraph of the open KG generated by \method-BERT$_{\rm LARGE}$ from the Wikipedia page ``Jacob\_van\_Ruisdael''. }
      \label{fig:kg4}}
\end{figure*}

\begin{figure*}
    \centering
    \includegraphics[width=0.5\textwidth]{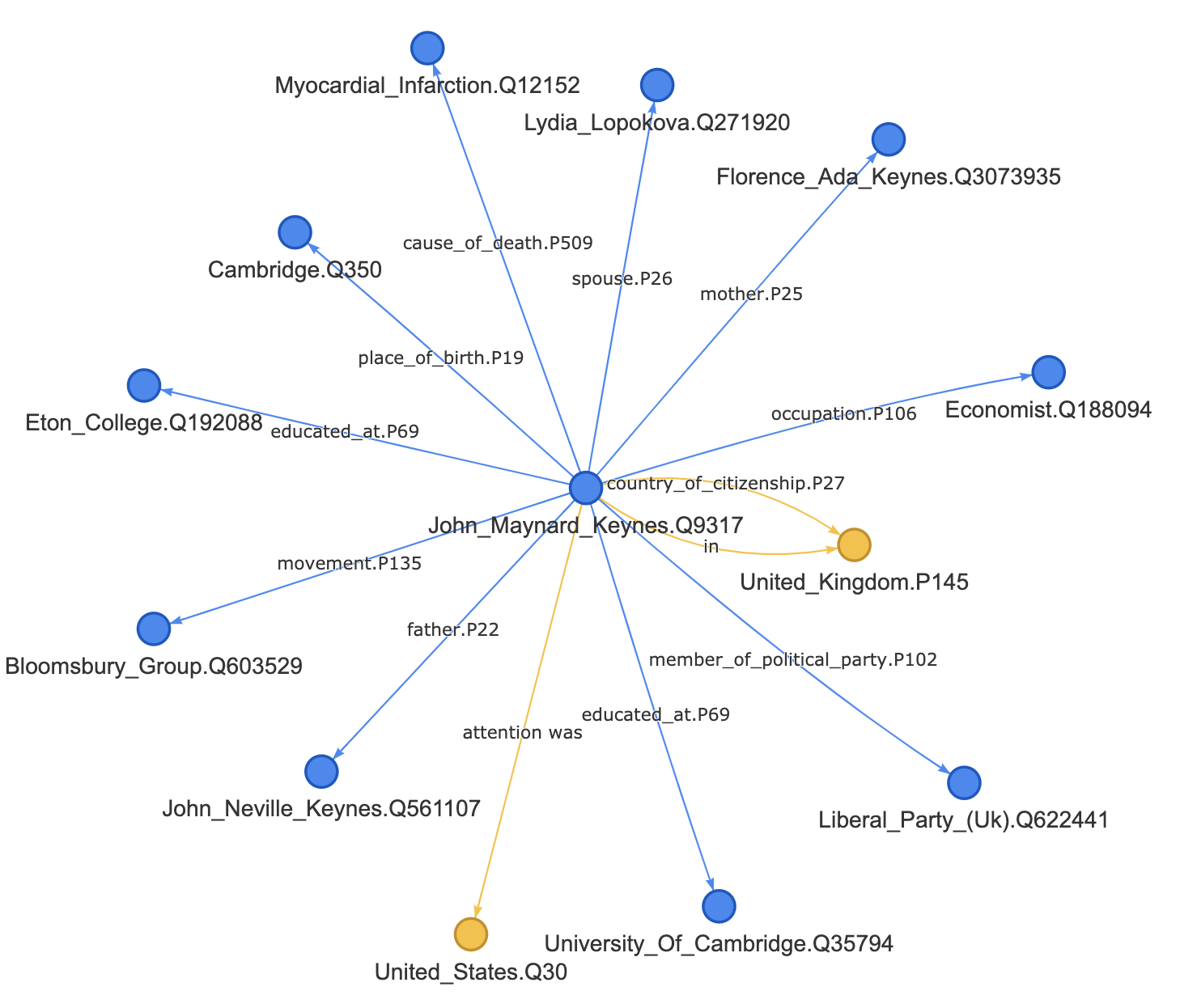}
    \caption{{\small A snapshot subgraph of the open KG generated by \method-BERT$_{\rm LARGE}$ from the Wikipedia page ``John\_Maynard\_Keynes''. }
      \label{fig:kg5}}
\end{figure*}

\begin{figure*}
    \centering
    \includegraphics[width=0.5\textwidth]{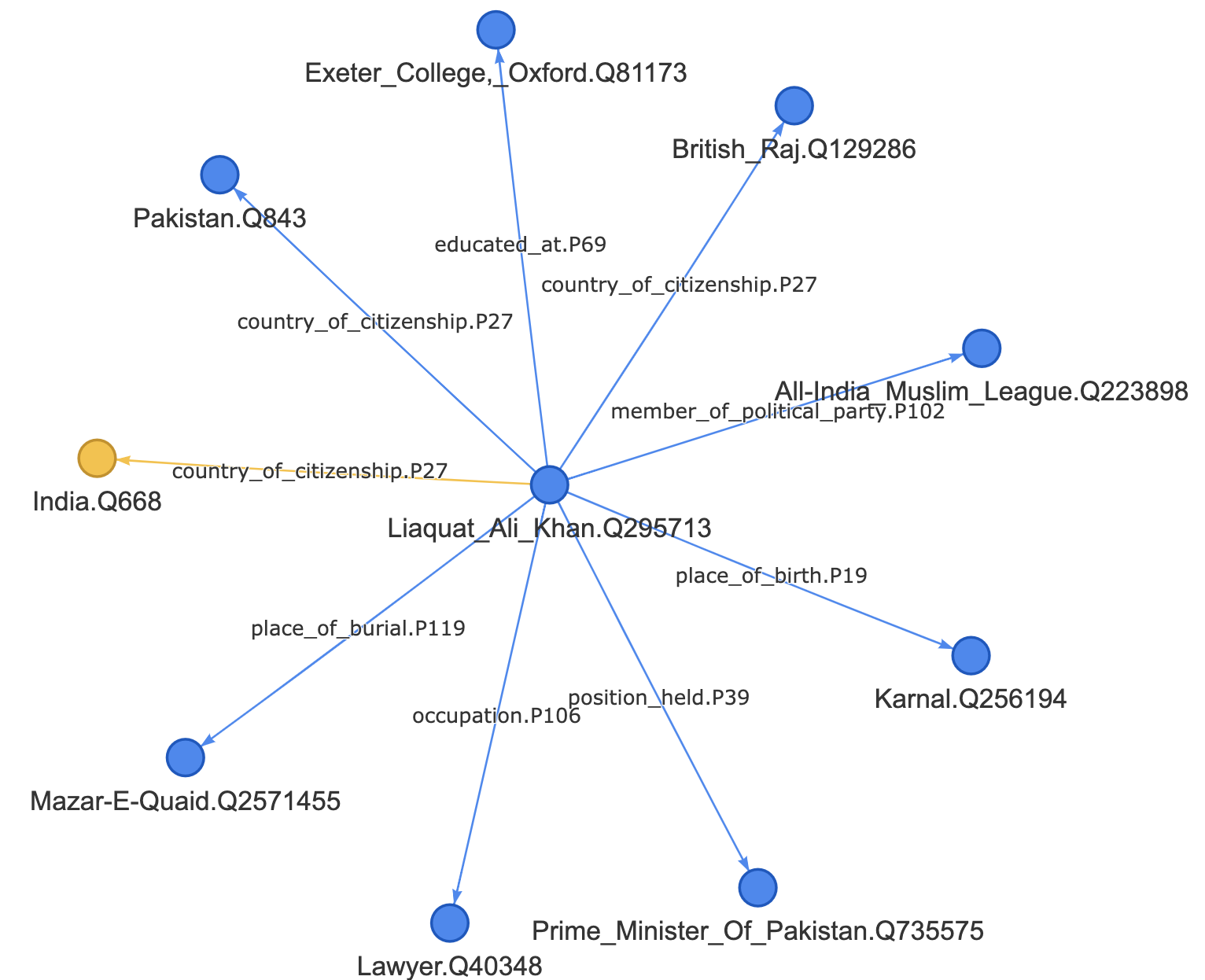}
    \caption{{\small A snapshot subgraph of the open KG generated by \method-BERT$_{\rm LARGE}$ from the Wikipedia page ``Liaquat\_Ali\_Khan''. }
      \label{fig:kg6}}
\end{figure*}

\begin{figure*}
    \centering
    \includegraphics[width=0.5\textwidth]{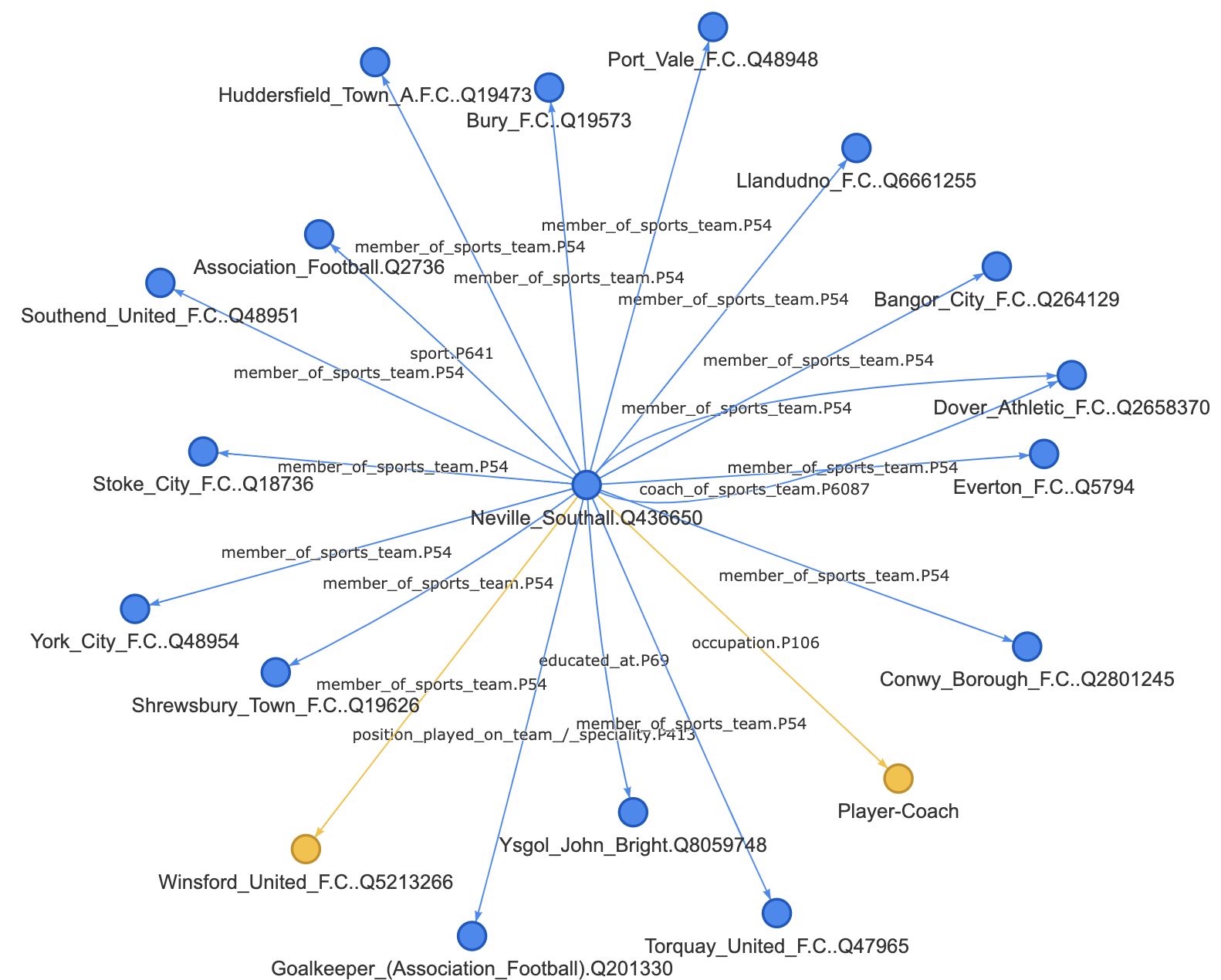}
    \caption{{\small A snapshot subgraph of the open KG generated by \method-BERT$_{\rm LARGE}$ from the Wikipedia page ``Neville\_Southall''. }
      \label{fig:kg7}}
\end{figure*}

\begin{figure*}
    \centering
    \includegraphics[width=0.5\textwidth]{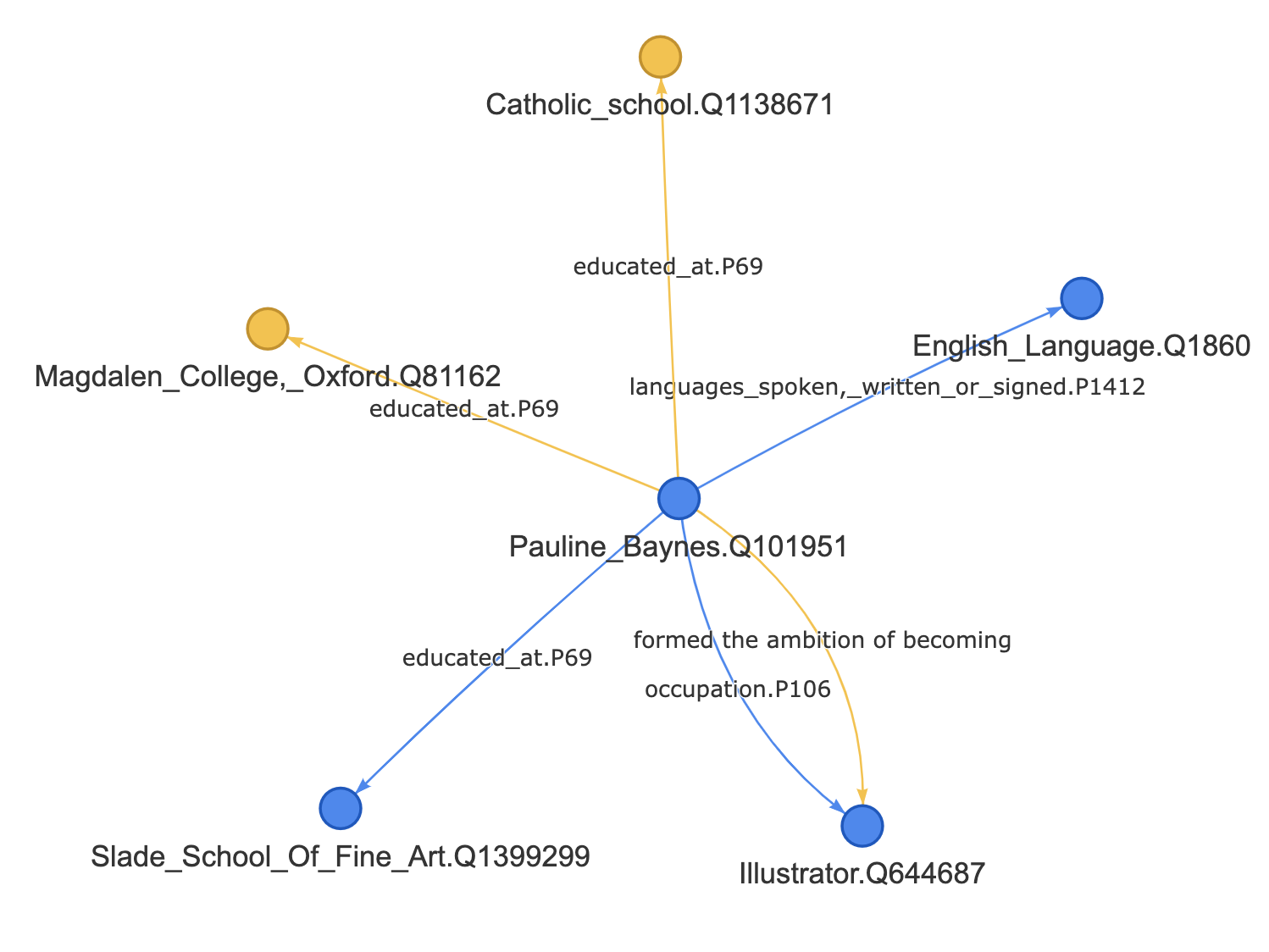}
    \caption{{\small A snapshot subgraph of the open KG generated by \method-BERT$_{\rm LARGE}$ from the Wikipedia page ``Pauline\_Baynes''. }
      \label{fig:kg9}}
\end{figure*}

\begin{figure*}
    \centering
    \includegraphics[width=0.5\textwidth]{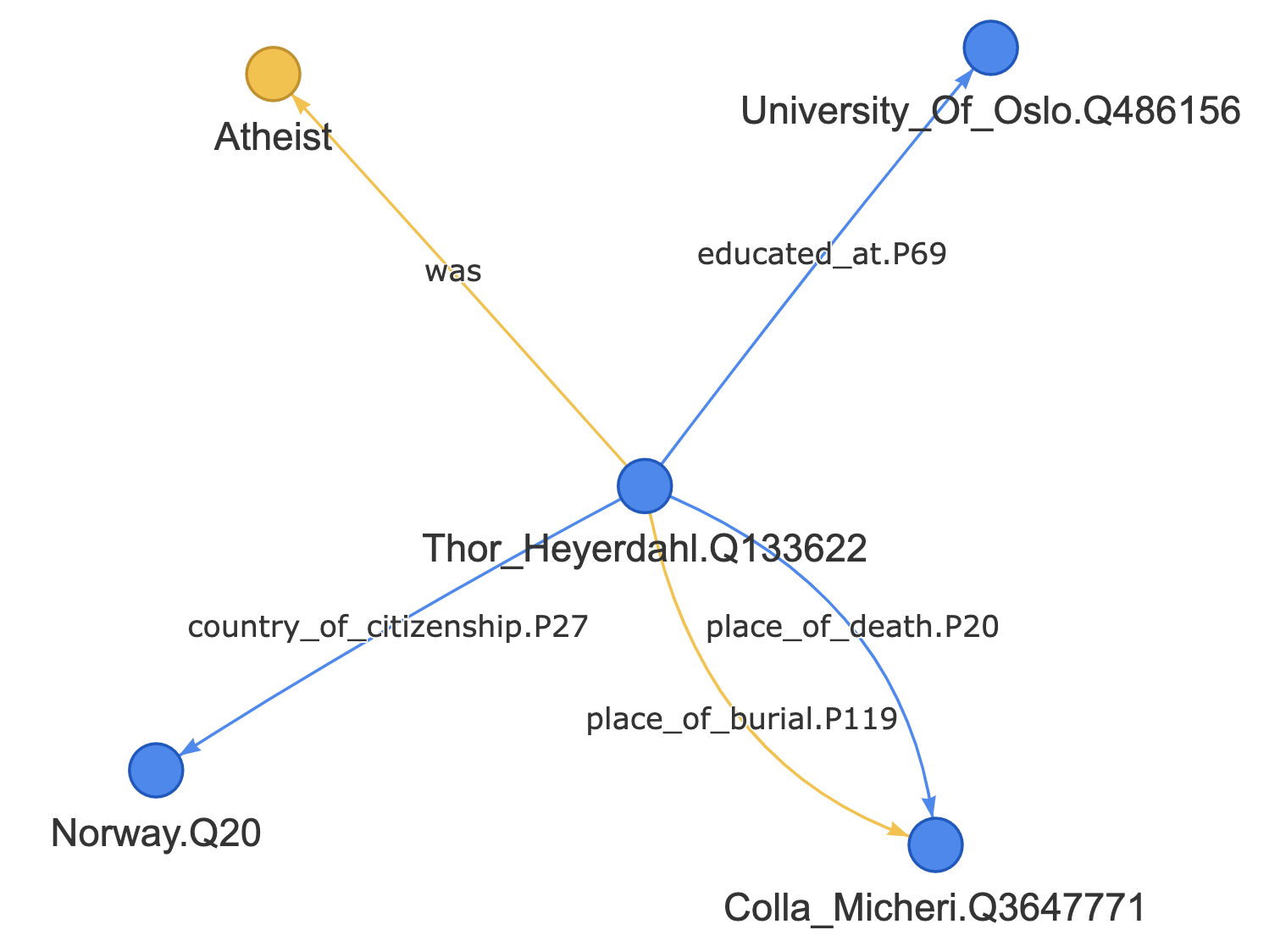}
    \caption{{\small A snapshot subgraph of the open KG generated by \method-BERT$_{\rm LARGE}$ from the Wikipedia page ``Thor\_Heyerdahl'. }
      \label{fig:kg10}}
\end{figure*}

\begin{figure*}
    \centering
    \includegraphics[width=0.5\textwidth]{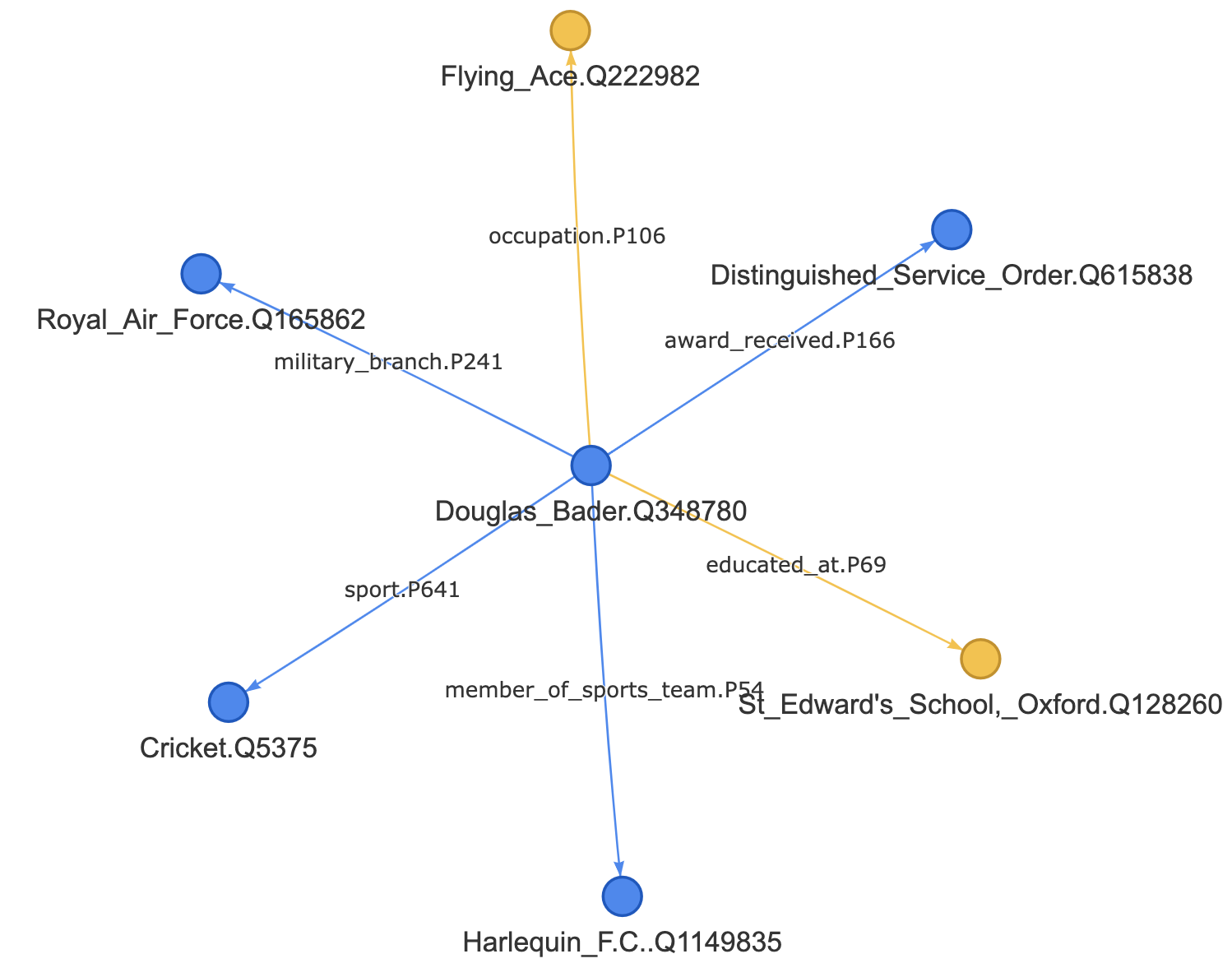}
    \caption{{\small A snapshot subgraph of the open KG generated by \method-GPT-2$_{\rm XL}$ from the Wikipedia page ``Douglas\_Bader''. }
      \label{fig:kg11}}
\end{figure*}

\begin{figure*}
    \centering
    \includegraphics[width=0.5\textwidth]{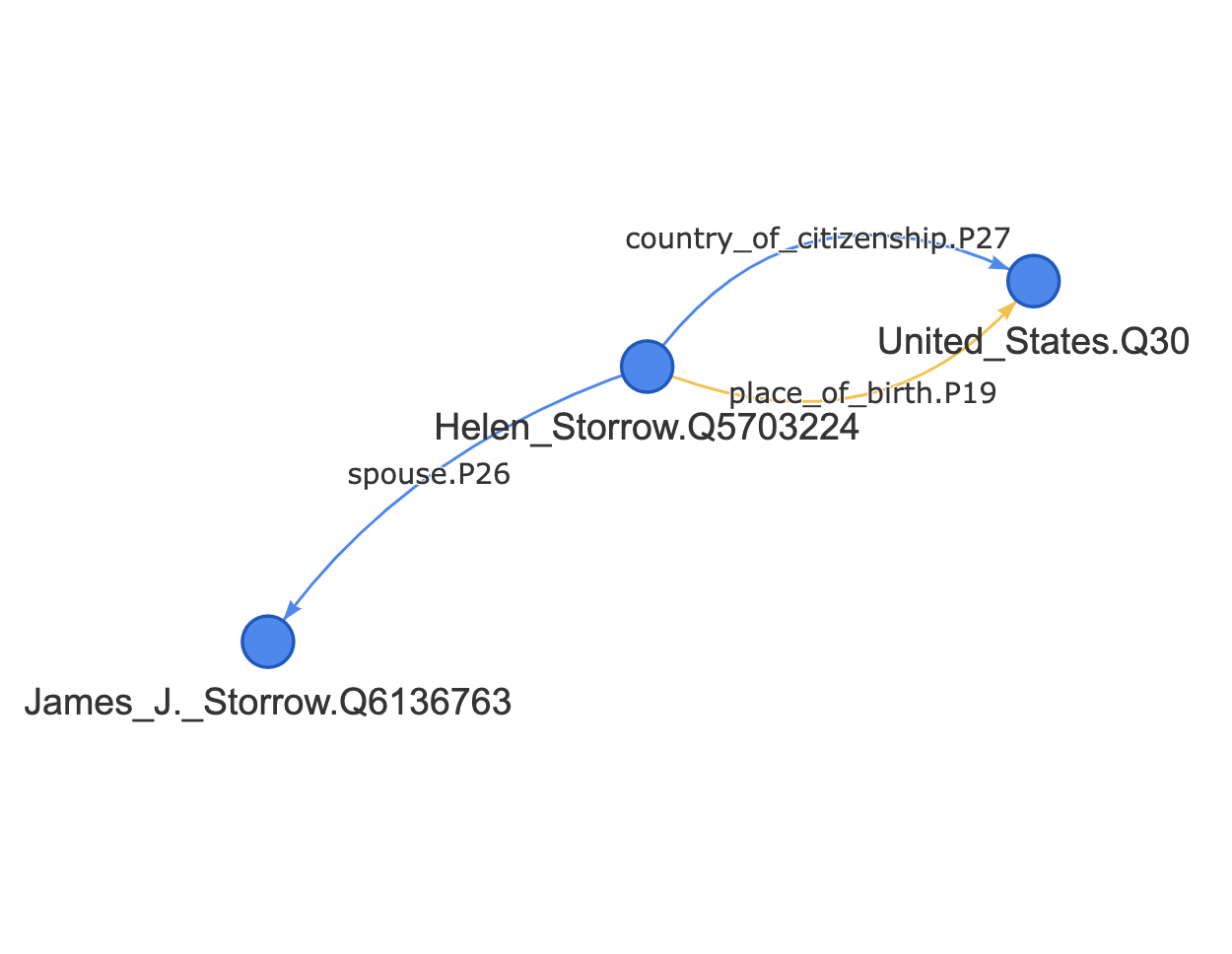}
    \caption{{\small A snapshot subgraph of the open KG generated by \method-GPT-2$_{\rm XL}$ from the Wikipedia page ``Helen\_Storrow''. }
      \label{fig:kg13}}
\end{figure*}

\begin{figure*}
    \centering
    \includegraphics[width=0.5\textwidth]{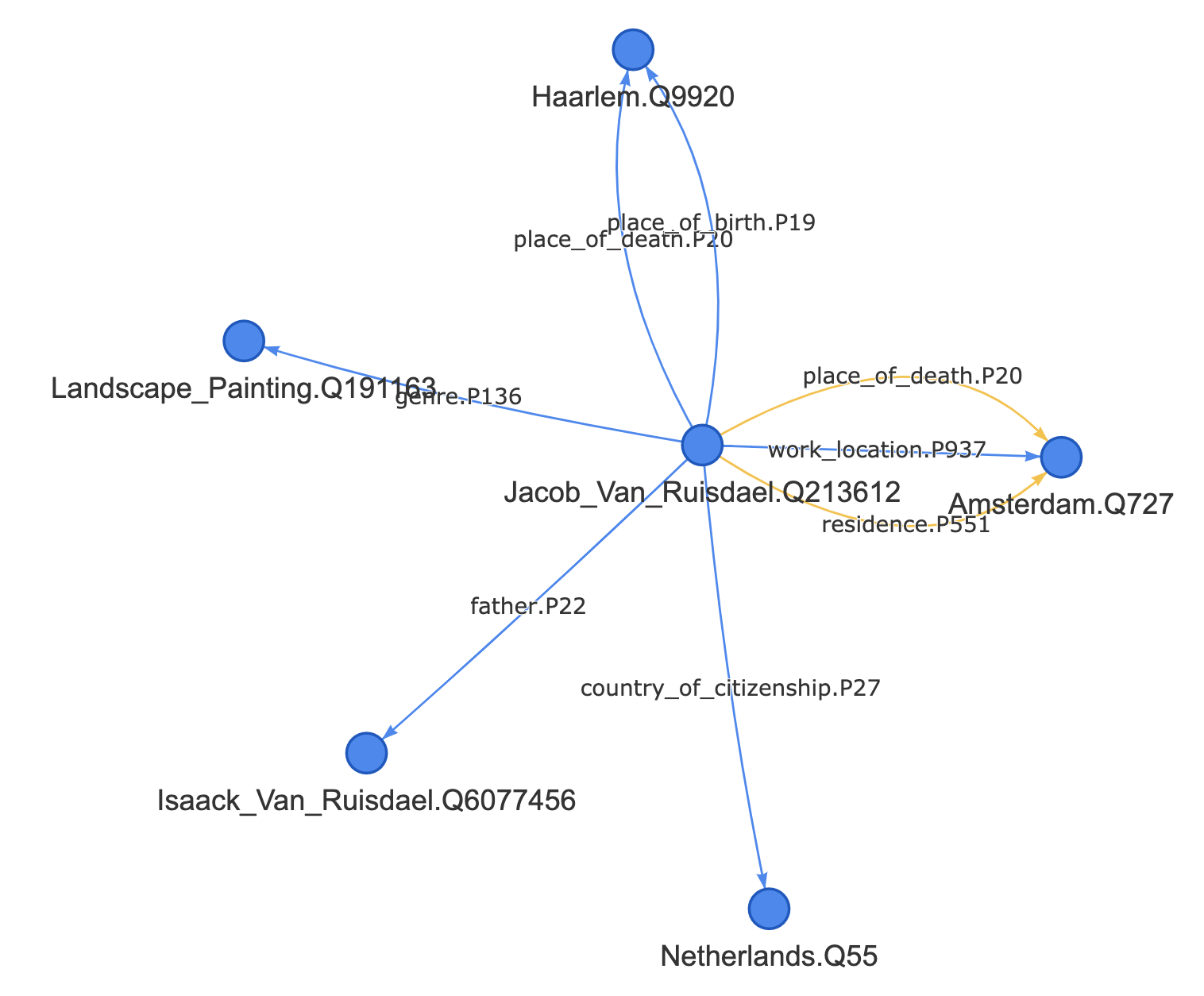}
    \caption{{\small A snapshot subgraph of the open KG generated by \method-GPT-2$_{\rm XL}$ from the Wikipedia page ``Jacob\_van\_Ruisdael''. }
      \label{fig:kg14}}
\end{figure*}

\begin{figure*}
    \centering
    \includegraphics[width=0.5\textwidth]{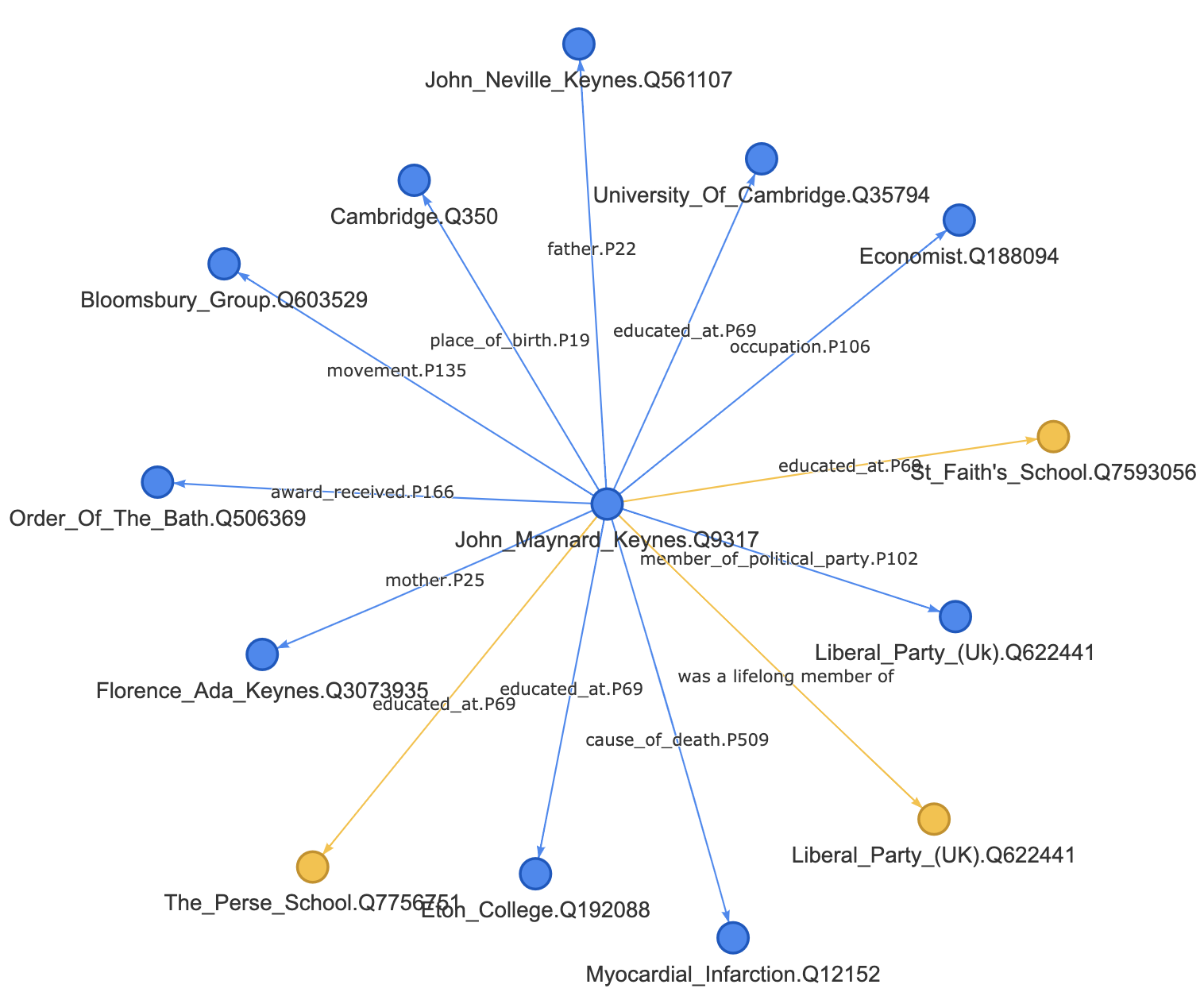}
    \caption{{\small A snapshot subgraph of the open KG generated by \method-GPT-2$_{\rm XL}$ from the Wikipedia page ``John\_Maynard\_Keynes''. }
      \label{fig:kg15}}
\end{figure*}

\begin{figure*}
    \centering
    \includegraphics[width=0.5\textwidth]{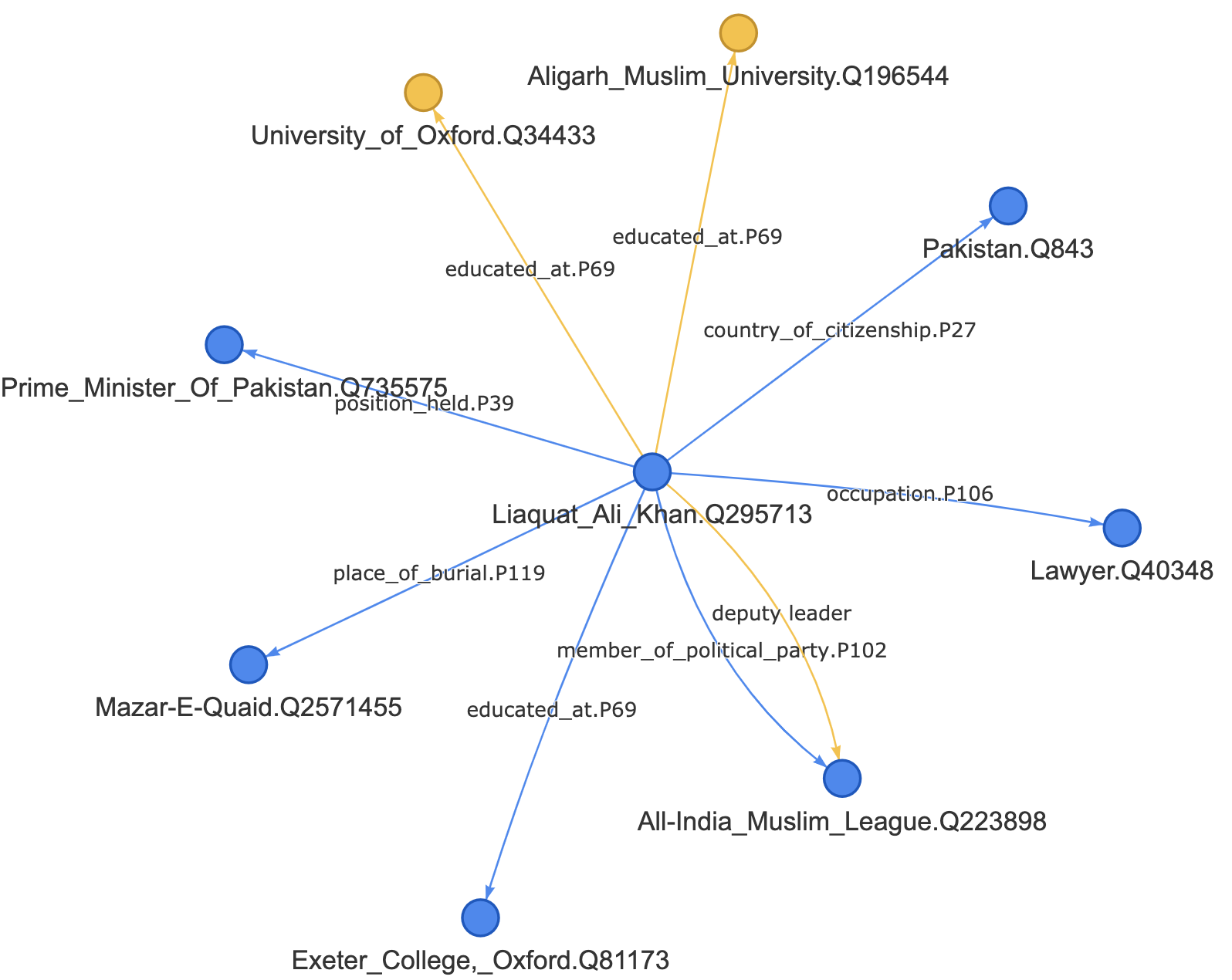}
    \caption{{\small A snapshot subgraph of the open KG generated by \method-GPT-2$_{\rm XL}$ from the Wikipedia page ``Liaquat\_Ali\_Khan''. }
      \label{fig:kg16}}
\end{figure*}

\begin{figure*}
    \centering
    \includegraphics[width=0.5\textwidth]{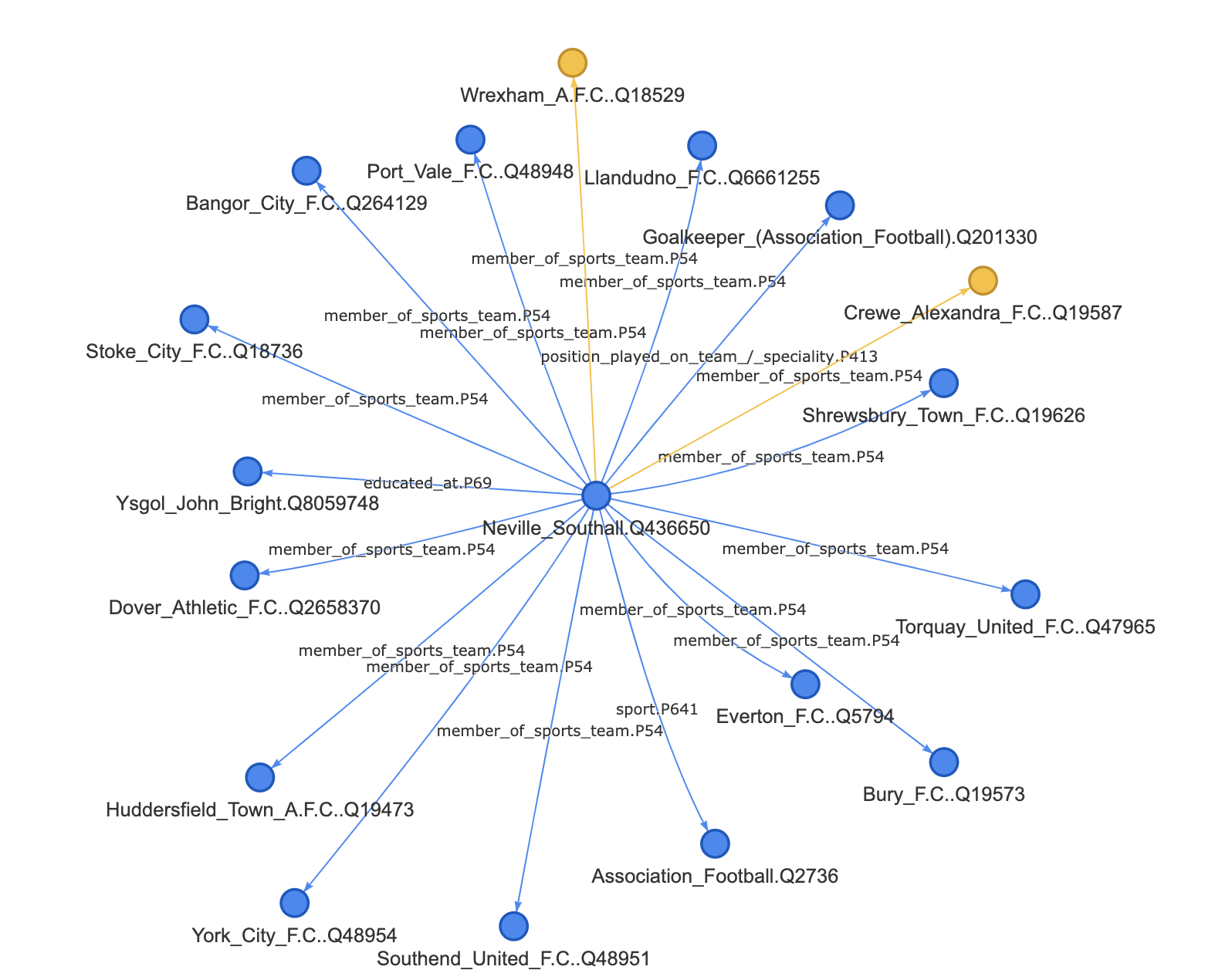}
    \caption{{\small A snapshot subgraph of the open KG generated by \method-GPT-2$_{\rm XL}$ from the Wikipedia page ``Neville\_Southall''. }
      \label{fig:kg17}}
\end{figure*}

\begin{figure*}
    \centering
    \includegraphics[width=0.5\textwidth]{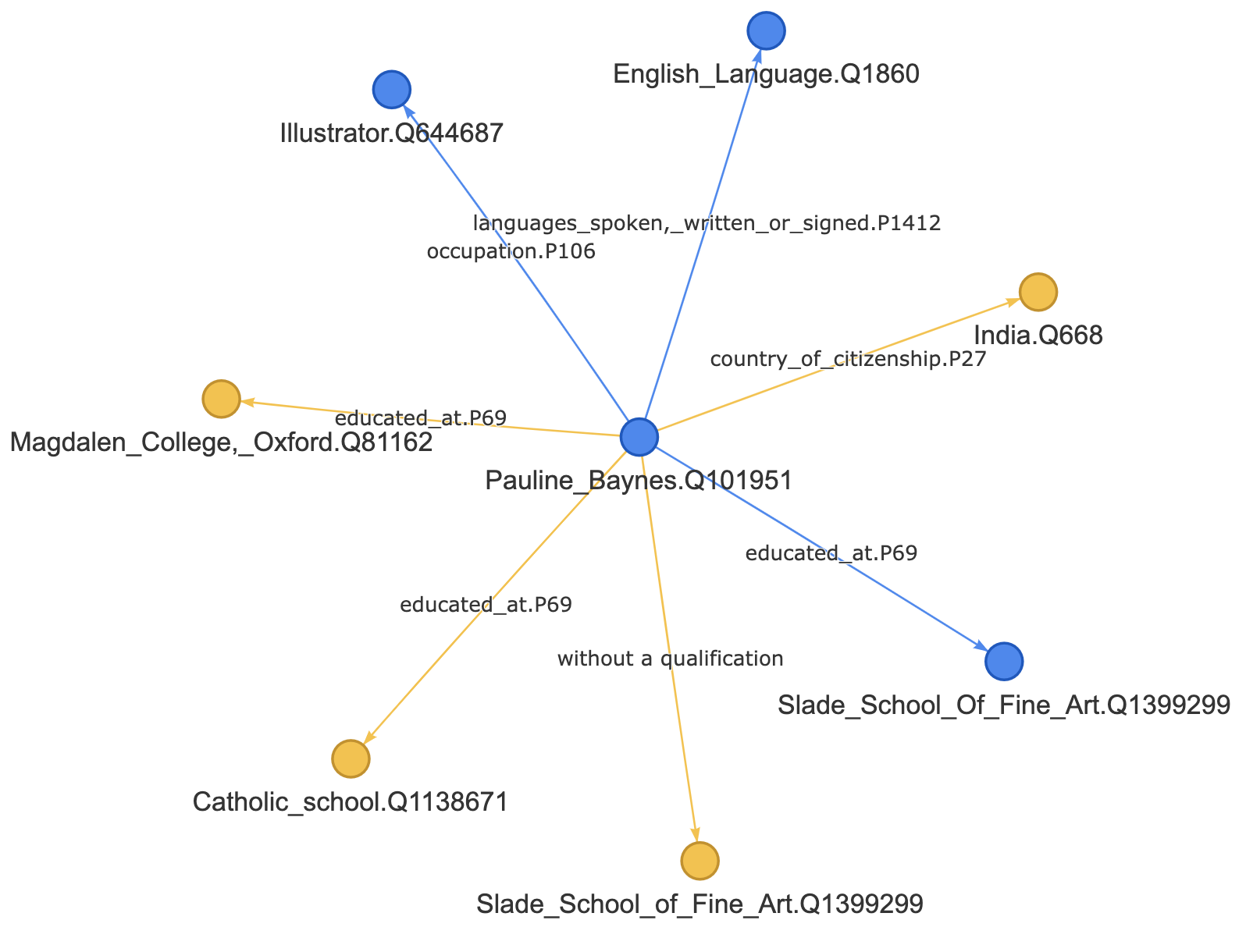}
    \caption{{\small A snapshot subgraph of the open KG generated by \method-GPT-2$_{\rm XL}$ from the Wikipedia page ``Pauline\_Baynes''. }
      \label{fig:kg19}}
\end{figure*}

\begin{figure*}
    \centering
    \includegraphics[width=0.35\textwidth]{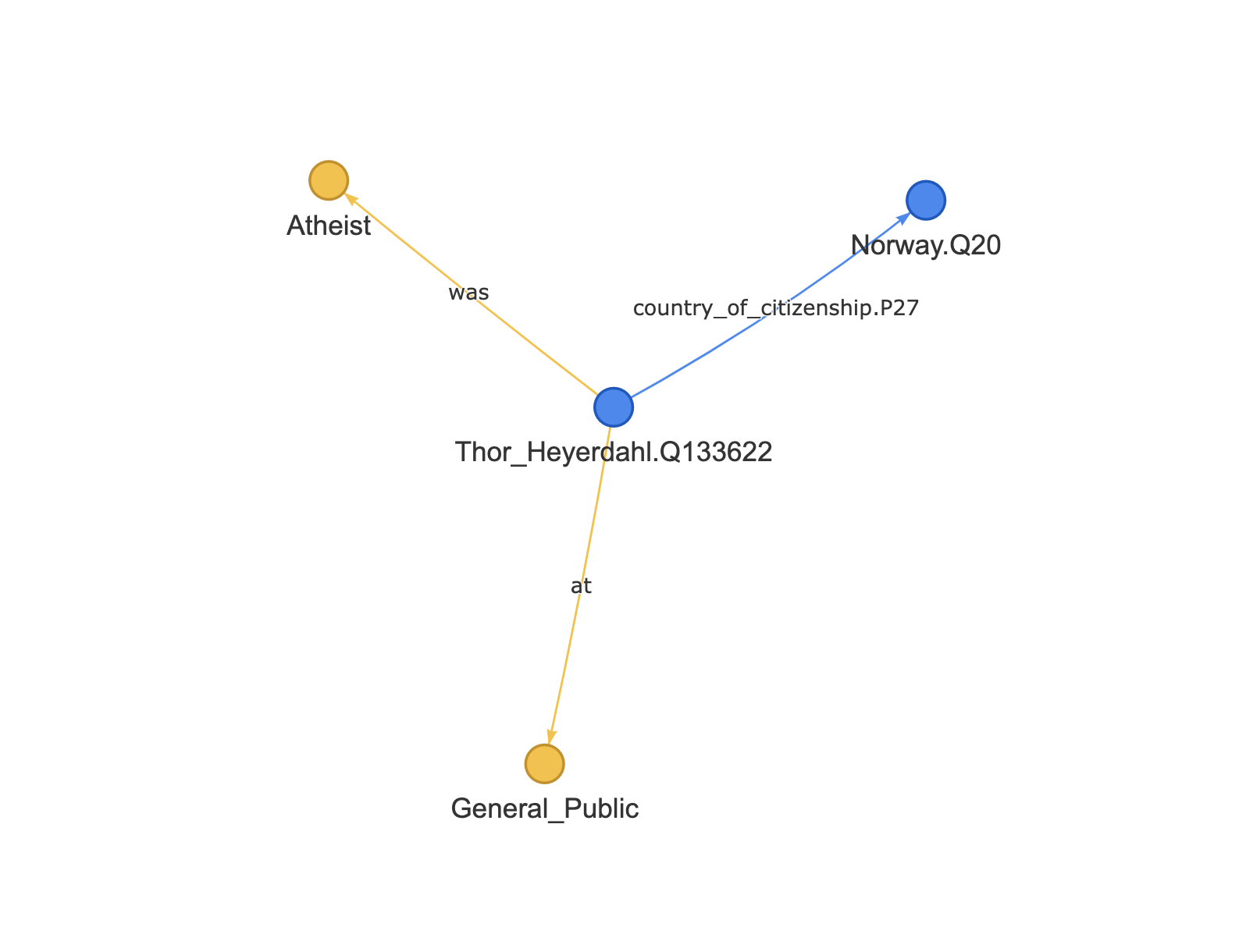}
    \caption{{\small A snapshot subgraph of the open KG generated by \method-GPT-2$_{\rm XL}$ from the Wikipedia page ``Thor\_Heyerdahl''. }
      \label{fig:kg20}}
\end{figure*}

We sample several documents from the Wikidata corpus. We visualize the mapped facts and unmapped facts from those documents as examples of subgraphs in the resulting open KGs. We show the snapshots of the subgraphs generated by \method-BERT$_{\rm LARGE}$ from Figure~\ref{fig:kg1} to Figure~\ref{fig:kg10}. We similarly illustrate the snapshots of the subgraphs constructed by \method-GPT-2$_{\rm XL}$ from Figure~\ref{fig:kg11} to Figure~\ref{fig:kg20}. In each figure, the blue node and arrow represent the mapped facts in the Wikidata schema, while the yellow node and arrow denote the unmapped facts in the open schema. We additionally visualize the correct facts that are new in Wikidata according to Sec.~\ref{sec:error} in yellow. 

\begin{figure}
\begin{center}
\tiny
\resizebox{0.99\linewidth}{!}
  {
\begin{tabular}{l p{11cm} p{11cm}}
\toprule
{\bf ID} & {\bf Sentence} & {\bf Candidate facts to mapped facts} \\ \hline
\input{appendices_table/mapped_bertlarge_kbp}
\bottomrule
\end{tabular}
}
\end{center}
\vspace{-0.1in}
\caption{{\small Mapped facts: \method-BERT$_{\rm LARGE}$ on TAC KBP.}} \label{fig:kbpmapbert}
\end{figure}

\begin{figure}
\begin{center}
\tiny
\resizebox{0.99\linewidth}{!}
  {
\begin{tabular}{l p{11cm} p{11cm}}
\toprule
{\bf ID} & {\bf Sentence} & {\bf Candidate facts to mapped facts} \\ \hline
\input{appendices_table/mapped_gpt2xl_kbp}
\bottomrule
\end{tabular}
}
\end{center}
\caption{{\small Mapped facts: \method-GPT-2$_{\rm XL}$ on TAC KBP.}} \label{fig:kbpmapgpt}
\end{figure}

\begin{figure}
\begin{center}
\tiny
\resizebox{1.0\linewidth}{!}
  {
\begin{tabular}{l p{7cm} p{7cm} p{7cm}}
\toprule
\multirow{2}{*}{\bf ID} & \multirow{2}{*}{\bf Sentence} & \multicolumn{2}{c}{\bf Candidate facts to mapped facts} \\
& & {\bf BERT$_{\rm LARGE}$} & {\bf GPT-2$_{\rm XL}$} \\ \hline
\input{appendices_table/mapped_compare_kbp}
\bottomrule
\end{tabular}
}
\end{center}
\caption{{\small Mapped facts: \method-BERT$_{\rm LARGE}$ vs. \method-GPT-2$_{\rm XL}$ on TAC KBP.}} \label{fig:kbpmapbertgpt}
\end{figure}

\begin{figure}
\begin{center}
\tiny
\resizebox{1.0\linewidth}{!}
  {
\begin{tabular}{l p{10cm} p{10cm}}
\toprule
{\bf ID} & {\bf Sentence} & {\bf Candidate facts to unmapped facts} \\ \hline
\input{appendices_table/unmapped_bertlarge_kbp}
\bottomrule
\end{tabular}
}
\end{center}
\caption{{\small Unmapped facts: \method-BERT$_{\rm LARGE}$ on TAC KBP.}} \label{fig:kbpunmapbert}
\end{figure}

\begin{figure}
\begin{center}
\tiny
\resizebox{1.0\linewidth}{!}
  {
\begin{tabular}{l p{10cm} p{10cm}}
\toprule
{\bf ID} & {\bf Sentence} & {\bf Candidate facts to unmapped facts} \\ \hline
\input{appendices_table/unmapped_gpt2xl_kbp}
\bottomrule
\end{tabular}
}
\end{center}
\caption{{\small Unmapped facts: \method-GPT-2$_{\rm XL}$ on TAC KBP.}} \label{fig:kbpunmapgpt}
\end{figure}

\begin{figure}
\begin{center}
\tiny
\resizebox{1.0\linewidth}{!}
  {
\begin{tabular}{l p{7cm} p{7cm} p{7cm}}
\toprule
\multirow{2}{*}{\bf ID} & \multirow{2}{*}{\bf Sentence} & \multicolumn{2}{c}{\bf Candidate facts to mapped facts} \\
& & {\bf BERT$_{\rm LARGE}$} & {\bf GPT-2$_{\rm XL}$} \\ \hline
\input{appendices_table/unmapped_compare_kbp}
\bottomrule
\end{tabular}
}
\end{center}
\caption{{\small Unmapped facts: \method-BERT$_{\rm LARGE}$ vs. \method-GPT-2$_{\rm XL}$ on TAC KBP.}} \label{fig:kbpunmapbertgpt}
\end{figure}

\begin{figure}
\begin{center}
\tiny
\resizebox{1.0\linewidth}{!}
  {
\begin{tabular}{l p{10cm} p{10cm}}
\toprule
{\bf ID} & {\bf Sentence} & {\bf Candidate facts to mapped facts} \\ \hline
\input{appendices_table/mapped_bertlarge_wikidata}
\bottomrule
\end{tabular}
}
\end{center}
\caption{{\small Mapped facts: \method-BERT$_{\rm LARGE}$ on Wikidata.}} \label{fig:wikimapbert}
\end{figure}

\begin{figure}
\begin{center}
\tiny
\resizebox{1.0\linewidth}{!}
  {
\begin{tabular}{l p{10cm} p{10cm}}
\toprule
{\bf ID} & {\bf Sentence} & {\bf Candidate facts to mapped facts} \\ \hline
\input{appendices_table/mapped_gpt2xl_wikidata}
\bottomrule
\end{tabular}
}
\end{center}
\caption{{\small Mapped facts: \method-GPT-2$_{\rm XL}$ on Wikidata.}} \label{fig:wikimapgpt}
\end{figure}

\begin{figure}
\begin{center}
\tiny
\resizebox{1.0\linewidth}{!}
  {
\begin{tabular}{l p{7cm} p{7cm} p{7cm}}
\toprule
\multirow{2}{*}{\bf ID} & \multirow{2}{*}{\bf Sentence} & \multicolumn{2}{c}{\bf Candidate facts to mapped facts} \\
& & {\bf BERT$_{\rm LARGE}$} & {\bf GPT-2$_{\rm XL}$} \\ \hline
\input{appendices_table/mapped_compare_wikidata}
\bottomrule
\end{tabular}
}
\end{center}
\caption{{\small Mapped facts: \method-BERT$_{\rm LARGE}$ vs. \method-GPT-2$_{\rm XL}$ on Wikidata.}} \label{fig:wikimapbertgpt}
\end{figure}

\begin{figure}
\begin{center}
\tiny
\resizebox{1.0\linewidth}{!}
  {
\begin{tabular}{l p{10cm} p{10cm}}
\toprule
{\bf ID} & {\bf Sentence} & {\bf Candidate facts to unmapped facts} \\ \hline
\input{appendices_table/unmapped_bertlarge_wikidata}
\bottomrule
\end{tabular}
}
\end{center}
\caption{{\small Unmapped facts: \method-BERT$_{\rm LARGE}$ on Wikidata.}} \label{fig:wikiunmapbert}
\end{figure}

\begin{figure}
\begin{center}
\tiny
\resizebox{1.0\linewidth}{!}
  {
\begin{tabular}{l p{10cm} p{10cm}}
\toprule
{\bf ID} & {\bf Sentence} & {\bf Candidate facts to unmapped facts} \\ \hline
\input{appendices_table/unmapped_gpt2xl_wikidata}
\bottomrule
\end{tabular}
}
\end{center}
\caption{{\small Unmapped facts: \method-GPT-2$_{\rm XL}$ on Wikidata.}} \label{fig:wikiunmapgpt}
\end{figure}

\begin{figure}
\begin{center}
\tiny
\resizebox{1.0\linewidth}{!}
  {
\begin{tabular}{l p{7cm} p{7cm} p{7cm}}
\toprule
\multirow{2}{*}{\bf ID} & \multirow{2}{*}{\bf Sentence} & \multicolumn{2}{c}{\bf Candidate facts to mapped facts} \\
& & {\bf BERT$_{\rm LARGE}$} & {\bf GPT-2$_{\rm XL}$} \\ \hline
\input{appendices_table/unmapped_compare_wikidata}
\bottomrule
\end{tabular}
}
\end{center}
\caption{{\small Unmapped facts: \method-BERT$_{\rm LARGE}$ vs. \method-GPT-2$_{\rm XL}$ on Wikidata.}} \label{fig:wikiunmapbertgpt}
\end{figure}

\end{document}